\documentclass{article}

\usepackage[,nonatbib,preprint]{neurips_2024}

\usepackage[utf8]{inputenc} 
\usepackage[T1]{fontenc}    
\usepackage{hyperref}       
\usepackage{url}            
\usepackage{booktabs}       
\usepackage{amsfonts}       
\usepackage{nicefrac}       
\usepackage{xcolor}   
\usepackage{graphicx}
\usepackage{makecell}
\usepackage{comment}
\usepackage{amsmath}
\usepackage{svg}

\usepackage{graphicx}
\usepackage{amsthm}
\usepackage{mathtools}
\usepackage{amssymb}
\usepackage[makeroom]{cancel}
\usepackage{tcolorbox}

\usepackage[config,font=small]{caption,subfig}
\usepackage{parskip}
\usepackage{enumerate}

\newcommand{\mat}[1]{\mathbf{#1}}

\title{KAN 2.0: \\ Kolmogorov-Arnold Networks Meet Science}

\author{%
Ziming Liu$^{1,4}$\thanks{zmliu@mit.edu} \quad Pingchuan Ma$^{1,3}$ \quad Yixuan Wang$^{2}$ \quad Wojciech Matusik$^{1,3}$ \quad Max Tegmark$^{1,4}$ \\
$^1$ Massachusetts Institute of Technology\\ $^2$ California Institute of Technology\\ $^3$ Computer Science and Artificial Intelligence Laboratory (CSAIL), MIT \\ $^4$ The NSF Institute for Artificial Intelligence and Fundamental Interactions
}

\begin{document}

\maketitle

\begin{abstract}
A major challenge of AI + Science lies in their inherent incompatibility: today's AI is primarily based on connectionism, while science depends on symbolism. To bridge the two worlds, we propose a framework to seamlessly synergize Kolmogorov-Arnold Networks (KANs) and science. The framework highlights KANs' usage for three aspects of scientific discovery: identifying relevant features, revealing modular structures, and discovering symbolic formulas. The synergy is bidirectional: science to KAN  (incorporating scientific knowledge into KANs), and KAN to science (extracting scientific insights from KANs). 
We highlight major new functionalities in \texttt{pykan}: 
(1) \textit{MultKAN}: KANs with multiplication nodes. (2) \textit{kanpiler}: a KAN compiler that compiles symbolic formulas into KANs. (3) \textit{tree converter}: convert KANs (or any neural networks) to tree graphs. Based on these tools, we demonstrate KANs' capability to discover various types of physical laws, including conserved quantities, Lagrangians, symmetries, and constitutive laws. 
\end{abstract}

\begin{figure}[hb]
    \centering
    \includegraphics[width=1.0\linewidth]{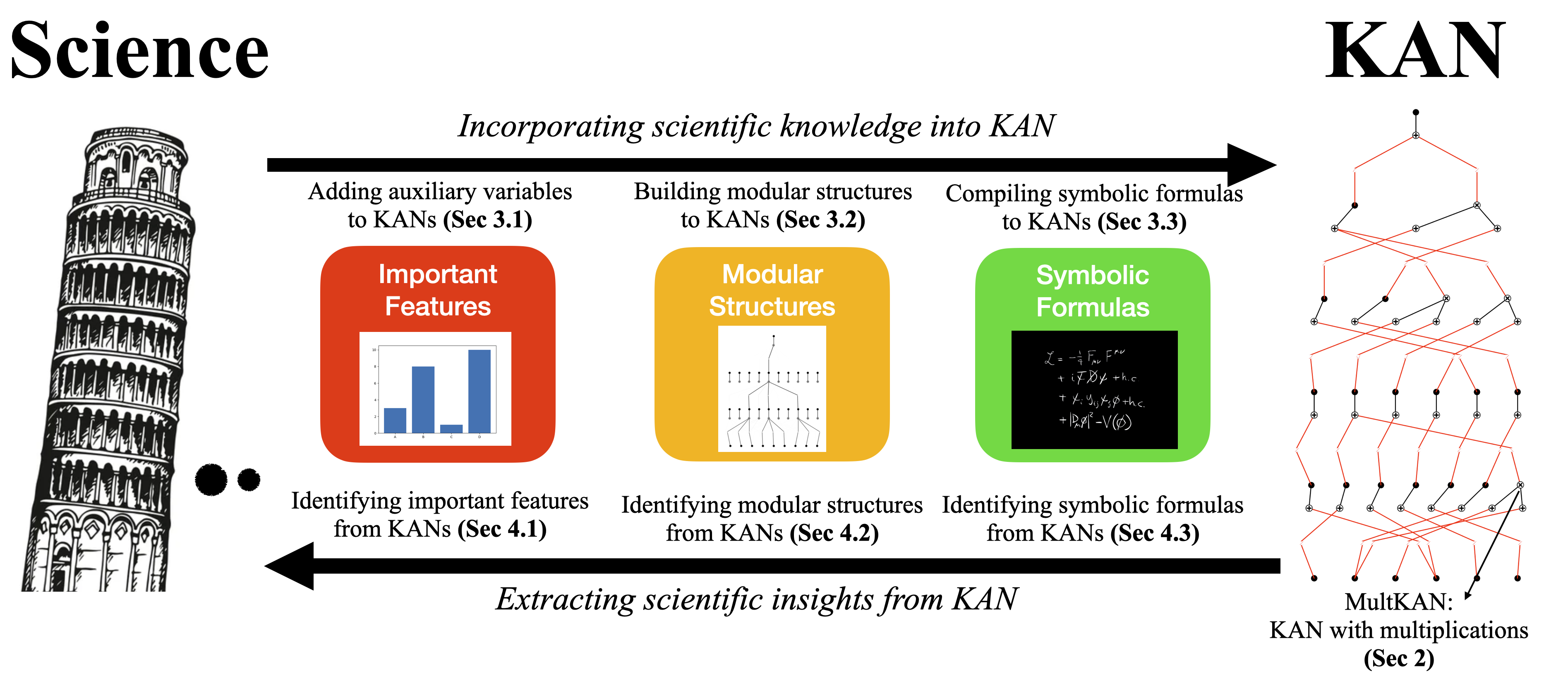}
    \caption{Synergizing science and the Kolmogorov-Arnold Network (KAN).}
    \label{fig:outline}
\end{figure}

\section{Introduction}
In recent years, AI + Science has emerged as a promising new field, leading to significant scientific advancements including protein folding prediction~\cite{jumper2021highly}, automated theorem proving~\cite{yang2024leandojo, trinh2024solving}, weather forecast~\cite{lam2023learning}, among others. A common thread among these tasks is that they can all be well formulated into problems with clear objectives, optimizable by black-box AI systems. While this paradigm works exceptionally well for application-driven science, a different kind of science exists:  curiosity-driven science. In curiosity-driven research, the procedure is more exploratory, often lacking clear goals beyond ``gaining more understanding''.  To clarify, curiosity-driven science is far from useless; quite the opposite. The scientific knowledge and understanding gained through curiosity often lay a solid foundation for tomorrow's technology and foster a wide range of applications. 

Although both application-driven and curiosity-driven science are invaluable and irreplaceable, they ask different questions. When astronomers observe the motion of celestial bodies, application-driven researchers focus on predicting their future states, while curiosity-driven researchers explore the physics behind the motion. Another example is AlphaFold, which, despite its tremendous success in predicting protein structures, remains in the realm of application-driven science because it does not provide new knowledge at a more fundamental level (e.g., atomic forces). Hypothetically, AlphaFold must have uncovered important unknown physics to achieve its highly accurate predictions. However, this information remains hidden from us, leaving AlphaFold largely a black box. Therefore, we advocate for new AI paradigms to support curiosity-driven science. This new paradigm of AI + Science demands a higher degree of interpretability and interactivity in AI tools so that they can be seamlessly integrated into scientific research. 

Recently, a new type of neural network called Kolmogorov-Arnold Network (KAN)~\cite{liu2024kan}, has shown promise for science-related tasks. Unlike multi-layer perceptrons (MLPs), which have fixed activation functions on nodes, KANs feature learnable activation functions on edges. Because KANs can decompose high-dimensional functions into one-dimensional functions, interpretability can be gained by symbolically regressing these 1D functions. However, their definition of interpretability is somewhat narrow, equating it almost exclusively with the ability to extract symbolic formulas. This limited definition restricts their scope, as symbolic formulas are not always necessary or feasible in science. For example, while symbolic equations are powerful and prevalent and physics, systems in chemistry and biology the systems are often too complex to be represented by such equations. In these fields, modular structures and key features may be sufficient to characterize interesting aspects of these systems. Another overlooked aspect is the reverse task of embedding knowledge into KANs: How can we incorporate prior knowledge into KANs, in the spirit of physics-informed learning? 

We enhance and extend KANs to make them easily used for curiosity-driven science. The goal of this paper can be summarized as follows:
\begin{center}
\begin{tcolorbox}[colframe=black, boxrule=1pt, colback=white, width=0.80\textwidth]
{\bf Goal}: Synergize Kolmogorov-Arnold Networks $\Leftrightarrow$ Science.

$\Leftarrow$: Build in scientific knowledge to KANs (Section 3).

$\Rightarrow$: Extract out scientific knowledge from KANs (Section 4).
\end{tcolorbox}
\end{center}

To be more concrete, scientific explanations may have different levels, ranging from the coarsest/easiest/correlational to the finest/hardest/causal: 
\begin{itemize}
    \item  Important features: For example, ``$y$ is fully determined by $x_1$ and $x_2$, while other factors do no matter.'' In other words, there exists a function $f$ such that $y=f(x_1,x_2)$.
    \item Modular structures: For instance, ``$x_1$ and $x_2$ contributes to $y$ independently in an additive way.`` This means there exists functions $g$ and $h$ such that $y=g(x_1)+h(x_2)$.
    \item Symbolic formulas: For example, ``$y$ depends on $x_1$ as a sine function and on $x_2$ as an exponential function''. In other words, $y={\rm sin}(x_1)+{\rm exp}(x_2)$.
\end{itemize}
The paper reports on how to incorporate and extract these properties from KANs. The structure of the paper is as follows (illustrated in Figure~\ref{fig:outline}): In Section 2, we augment the original KAN with multiplication nodes, introducing a new model called MultKAN. In Section 3, we explore ways to embed scientific inductive biases into KANs, focusing on important features (Section 3.1), modular structures (Section 3.2), and symbolic formulas (Section 3.3). In Section 4, we propose methods to extract scientific knowledge from KANs, again covering important features (Section 4.1), modular structures (Section 4.2), and symbolic formulas (Section 4.3).
In Section 5, we apply KANs to various scientific discovery tasks using the tools developed in the previous sections. These tasks include discovering conserved quantities, symmetries, Lagrangians, and constitutive laws. 
Codes are available at \url{https://github.com/KindXiaoming/pykan} and can also be installed via \texttt{pip install pykan}. Although the title of the paper is ``KAN 2.0'', the release version of \texttt{pykan} is 0.2.x.

\section{MultKAN: Augmenting KANs with multiplications}

\begin{figure}
    \centering
    \includegraphics[width=1.0\linewidth]{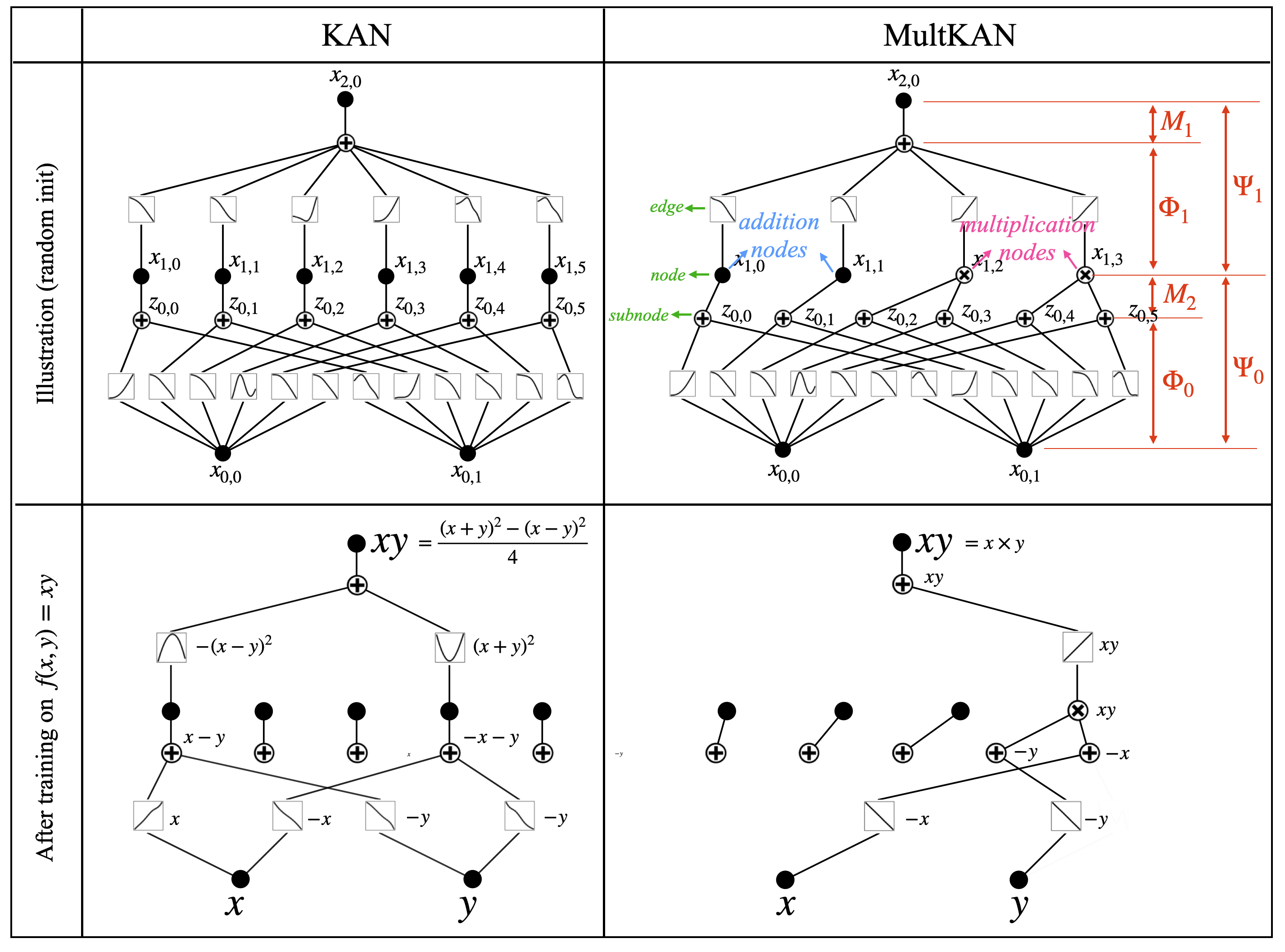}
    \caption{Top: comparing KAN and MultKAN diagrams. MultKAN has extra multiplication layers $\mat{M}$. Bottom: After training on $f(x,y)=xy$, KAN learns an algorithm requiring two addition nodes, while MultKAN requires only one multiplication node.}
    \label{fig:multkan-diagram}
\end{figure}

The Kolmogorov-Arnold representation theorem (KART) states that any continuous high-dimensional function can be decomposed into a finite composition of univariate continuous functions and additions:
\begin{equation}\label{eq:KART}
    f(\mat{x}) = f(x_1,\cdots,x_n)=\sum_{q=1}^{2n+1} \Phi_q\left(\sum_{p=1}^n\phi_{q,p}(x_p)\right).
\end{equation}
This implies that addition is the only true multivariate operation, while other multivariate operations (including multiplication) can be expressed as additions combined with univariate functions. For example, to multiply two positive numbers $x$ and $y$, we can express this as $xy={\rm exp}({\rm log}x+{\rm log}y)$~\footnote{If $x$ and $y$ can be negative, one may choose a large $c>0$ and express $xy={\rm exp}({\rm log}(x+c)+{\rm log}(y+c))-c(x+y)-c^2$. Other constructions include quadratic functions, such as $xy=((x+y)^2-(x-y)^2)/4$ or $xy=((x+y)^2-x^2-y^2)/2$.} whose right-hand side only consists of addition and univariate functions (log and exp).

However, given the prevalence of multiplications in both science and everyday life, it is desirable to explicitly include multiplications in KANs, which could potentially enhance both interpretability and capacity.

{\bf Kolmogorov-Arnold Network (KAN)} While the KART Eq.~(\ref{eq:KART}) corresponds to a two-layer network, Liu et al.~\cite{liu2024kan} managed to extend it to arbitrary depths by recognizing that seemingly different outer functions $\Phi_q$ and inner functions $\phi_{q,p}$ can be unified through their proposed \textit{KAN layers}. A depth-$L$ KAN can be constructed simply by stacking $L$ KAN layers. The shape of a depth-$L$ KAN is represented by an integer array $[n_0,n_1,\cdots,n_L]$ where $n_l$ denotes the number of neurons in the $l^{\rm th}$ neuron layers. The $l^{\rm th}$ KAN layer, with $n_l$ input dimensions and $n_{l+1}$ output dimensions, transforms an input vector $\mat{x}_l\in\mathbb{R}^{n_l}$ to $\mat{x}_{l+1}\in\mathbb{R}^{n_{l+1}}$
\begin{equation}\label{eq:kanforwardmatrix}
    \mat{x}_{l+1} = 
    \underbrace{\begin{pmatrix}
        \phi_{l,1,1}(\cdot) & \phi_{l,2,1}(\cdot) & \cdots & \phi_{l,n_{l},1}(\cdot) \\
        \phi_{l,1,2}(\cdot) & \phi_{l,2,2}(\cdot) & \cdots & \phi_{l,n_{l},2}(\cdot) \\
        \vdots & \vdots & & \vdots \\
        \phi_{l,1,n_{l+1}}(\cdot) & \phi_{l,2,n_{l+1}}(\cdot) & \cdots & \phi_{l,n_l,n_{l+1}}(\cdot) \\
    \end{pmatrix}}_{\mat{\Phi}_l}
    \mat{x}_{l},
\end{equation}
and the whole network is a composition of  $L$ KAN layers, i.e.,
\begin{equation}
    {\rm KAN}(\mat{x})=(\mat{\Phi}_{L-1}\circ\cdots\circ\mat{\Phi}_1\circ\mat{\Phi}_0)\mat{x}.
\end{equation}
In diagrams, KANs can be intuitively visualized as a network consisting of \textit{nodes} (summation) and \textit{edges} (learnable activations), as shown in Figure~\ref{fig:multkan-diagram} top left. When trained on the dataset generated from $f(x,y)=xy$, the KAN (Figure~\ref{fig:multkan-diagram} bottom left) uses two addition nodes, making it unclear what the network is doing. However, after some consideration, we realize it leverages the equality $xy = ((x+y)^2-(x-y)^2)/4$ but this is far from obvious.

{\bf Multiplicative Kolmogorov-Arnold Networks (MultKAN)} To explicitly introduce multiplication operations, we propose the MultKAN, which can reveal multiplicative structures in data more clearly. A MultKAN (shown in Figure~\ref{fig:multkan-diagram} top right) is similar to a KAN, with both having standard KAN layers. We refer to the input nodes of a KAN layer as \textit{nodes}, and the output nodes of a KAN layer \textit{subnodes}. The difference between KAN and MultKAN lies in the transformations from the current layer's subnodes to the next layer's nodes. In KANs, nodes are directly copied from the previous layer's subnodes. In MultKANs, some nodes (\textit{addition nodes}) are copied from corresponding subnodes, while other nodes (\textit{multiplication nodes}) perform multiplication on $k$ subnodes from the previous layer. For simplicity, we set $k=2$ below~\footnote{We set $k=2$ for simplicity, but the \texttt{pykan} package allows $k$ to be any integer $k\geq 2$. Users can even set different $k$ values for different multiplication nodes. However, if different $k$s values are used within the same layer, it can be challenging to parallelize these multiplications.}. 

Based on the MultKAN diagram (Figure~\ref{fig:multkan-diagram} top right), it can be intuitively understood that a MultKAN is a normal KAN with optional multiplications inserted in. To be mathematically precise, we define the following notations: The number of addition (multiplication) operations in layer $l$ are denoted as $n_l^a$ ($n_l^m$), respectively. These are collected into arrays: addition width $\mat{n}^a\equiv [n^a_0,n^a_1,\cdots,n^a_L]$ and multiplication width $\mat{n}^m\equiv [n^m_0,n^m_1,\cdots,n^m_L]$. When $n^m_0=n^m_1=\cdots=n^m_L=0$, the MultKAN reduces to a KAN. For example,   Figure~\ref{fig:multkan-diagram} (top right) shows a MultKAN with $\mat{n}^a=[2,2,1]$ and $\mat{n}^m=[0,2,0]$.

A MultKAN layer consists of a standard KANLayer $\mat{\Phi}_l$ and a multiplication layer $\mat{M}_l$. $\mat{\Phi}_l$ takes in an input vector $\mat{x}_l\in\mathbb{R}^{n_l^a+n_l^m}$ and outputs  $\mat{z}_l=\mat{\Phi}_l(\mat{x})\in\mathbb{R}^{n_{l+1}^a+2n_{l+1}^m}$. The multiplication layer consists of two parts: the multiplication part performs multiplications on subnode pairs, while the other part performs identity transformation.  Written in Python, $\mat{M}_l$ transforms $\mat{z}_l$ as follows:
\begin{equation}
    \mat{M}_l(\mat{z}_l) = {\rm concatenate}(\mat{z}_l[:n_{l+1}^a], \mat{z}_l[n_{l+1}^a::2] \odot \mat{z}_l[n_{l+1}^a+1::2])\in\mathbb{R}^{n_{l+1}^a+n_{l+1}^m},\end{equation}
where $\odot$ is element-wise multiplication. The MultKANLayer can be succinctly represented as $\mat{\Psi}_l\equiv \mat{M}_l\circ \mat{\Phi}_l$. The whole MultKAN is thus:
\begin{equation}
    {\rm MultKAN}(\mat{x})=(\mat{\Psi}_L\circ\mat{\Psi}_{L-1}\circ\cdots\circ\mat{\Psi}_1\circ\mat{\Psi}_0)\mat{x}.
\end{equation}
Since there are no trainable parameters in multiplication layers, all sparse regularization techniques (e.g., $\ell_1$ and entropy regularization) for KANs~\cite{liu2024kan} can be directly applied to MultKANs. For the multiplication task $f(x,y)=xy$, the MultKAN indeed learns to use one multiplication node, making it perform simple multiplication, as all the learned activation functions are linear (Figure~\ref{fig:multkan-diagram} bottom right). 

Although KANs have previously been seen as a special case of MultKANs, we extend the definition and treat ``KAN'' and ``MultKAN'' as synonyms. By default, when we refer to KANs, multiplication is allowed. If we specifically refer to a KAN without multiplication, we will explicitly state so.  

\section{Science to KANs}

In science, domain knowledge is crucial, allowing us to work effectively even with small or zero data. Therefore, it is beneficial to adopt a physics-informed approach for KANs: we should incorporate available inductive biases into KANs while preserving their flexibility to discover new physics from data.

We explore three types of inductive biases that can be integrated into KANs. From the coarsest/easiest/correlational to the finest/hardest/causal, they are important features (Section 3.1), modular structures (Section 3.2) and symbolic formulas (Section 3.3). 

\subsection{Adding important features to KANs}

In a regression problem, the goal is to find a function $f$ such that $y=f(x_1,x_2,\cdots, x_n)$. Suppose we want to introduce an auxiliary input variable $a = a(x_1,x_2,\dots, x_n)$, transforming the function to $y=f(x_1,\cdots, x_n, a)$. Although the auxiliary variable $a$ does not add new information, it can increase the expressive power of the neural network. This is because the network does not need to expend resources to calculate the auxiliary variable. Additionally, the computations may become simpler, leading to improved interpretability. Users can add auxiliary features to inputs using  the \texttt{augment\_input} method:
\begin{equation}
\texttt{model.augment\_input(original\_variables, auxiliary\_variables, dataset)}
\end{equation}

\begin{figure}
    \centering
    \includegraphics[width=1.0\linewidth]{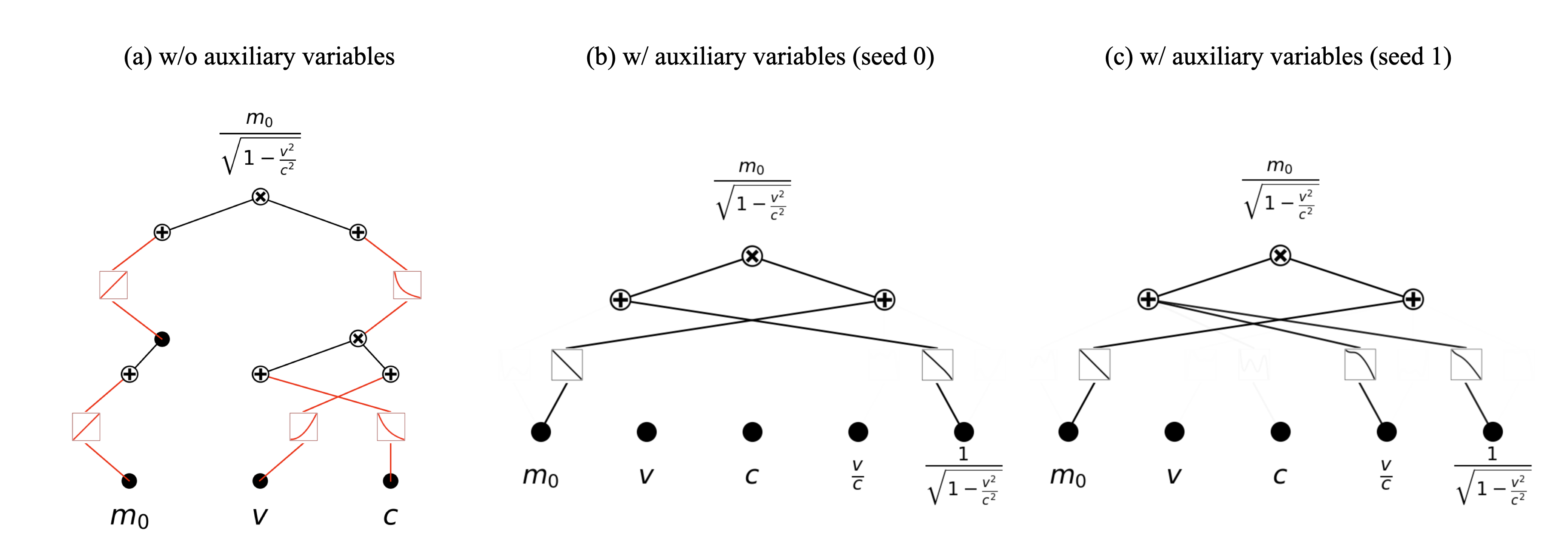}
    \caption{Adding auxiliary variables to inputs enhances interpretability. For the relativistic mass equation, $m=m_0/\sqrt{1-v^2/c^2}$, (a) a two-layer KAN is needed if only $(m_0, v, c)$ are used as inputs. (b) If we add $\beta\equiv v/c$ and $\gamma\equiv1/\sqrt{1-\beta^2}$ as auxiliary variables to KANs, a one-layer KAN suffices (seed 0). (c) seed 1 finds a different solution, which is sub-optimal and can be avoided through hypothesis testing (Section 4.3).}
    \label{fig:aux_var}
\end{figure}
As an example, consider the formula for relativistic mass $m(m_0,v,c)=m_0/\sqrt{1-(v/c)^2}$ where $m_0$ is the rest mass, $v$ is the velocity of the point mass, and $c$ is the speed of light. Since physicists often work with dimensionless numbers $\beta\equiv v/c$  and $\gamma\equiv 1/\sqrt{1-\beta^2}\equiv 1/\sqrt{1-(v/c)^2}$, they might introduce $\beta$ and $\gamma$ alongside $v$ and $c$ as inputs. Figure~\ref{fig:aux_var}, shows KANs with and without these auxiliary variables: (a) illustrates the KAN compiled from the symbolic formula (see Section 3.3 for the KAN compiler), which requires 5 edges; (b)(c) shows KANs with auxiliary variables, requiring only 2 or 3 edges and achieving loses of $10^{-6}$ and $10^{-4}$, respectively. Note that (b) and (c) differ only in random seeds. Seed 1 represents a sub-optimal solution because it also identifies $\beta=v/c$ as a key feature. This is not surprising, as in the classical limit $v\ll c$, $\gamma\equiv 1/\sqrt{1-(v/c)^2}\approx 1 + (v/c)^2/2=1+\beta^2/2$. 
The variation due to different seeds can be seen either as a feature or a bug: As a feature, this diversity can help find sub-optimal solutions which may nevertheless offer interesting insights; as a bug, it can be eliminated using the hypothesis testing method proposed in Section 4.3.

\subsection{Building modular structures to KANs}

\begin{figure}
    \centering
    \includegraphics[width=1.0\linewidth]{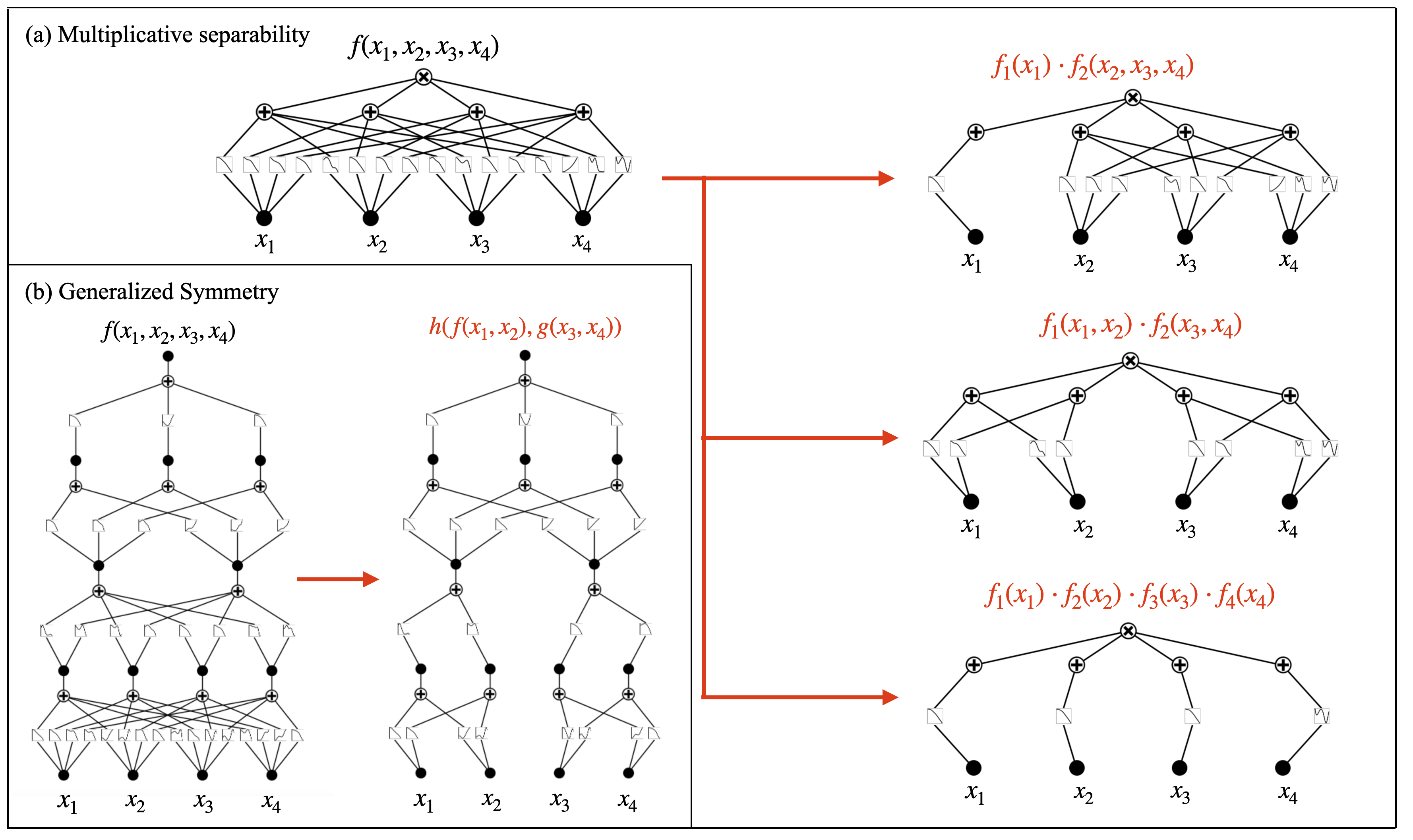}
    \caption{Building modular structures to KANs: (a) multiplicative separability;(b) symmetries.}
    \label{fig:build_structure}
\end{figure}

Modularity is prevalent in nature: for example, the human cerebral cortex is divided into several functionally distinct modules, each of these modules responsible for specific tasks such as perception or decision making. This modularity simplifies the understanding of neural networks, as it allows us to interpret clusters of neurons collectively rather than analyzing each neuron individually. Structural modularity is characterized by clusters of connections where intra-cluster connections are much stronger than inter-cluster ones. To enforce modularity, we introduce the \texttt{module} method, which preserves intra-cluster connections while removing inter-cluster connections. The modules are specified by users. The syntax is 
\begin{equation}
\texttt{model.module(start\_layer\_id, `[nodes\_id]->[subnodes\_id]->[nodes\_id]...'})  
\end{equation}
For example, if a user wants to assign specific nodes/subnodes to a module -- say, the $0^{\rm th}$ node in layer 1, the $1^{\rm st}$ and $3^{\rm rd}$ subnode in layer 1, the $1^{\rm st}$ and $3^{\rm rd}$ node in layer 2 --  they might use $\texttt{module(1,`[0]->[1,3]->[1,3]')}$. To be concrete, there are two types of modularity: separability and symmetry. 

{\bf Separability} 
We say a function is considered separable if it can be expressed as a sum or product of functions of non-overlapping variable groups. For example, a four-variable function $f(x_1,x_2,x_3,x_4)$
is maximally multiplicatively separable if it has the form $f_1(x_1)f_2(x_2)f_3(x_3)f_4(x_4)$, creating four distinct groups $(1),(2),(3),(4)$. Users can create these modules by calling the \texttt{module} method four times: \texttt{module(0,`[i]->[i]')}, $i=0,1,2,3$, shown in Figure~\ref{fig:build_structure} (a). The final call may be skipped since the first three are sufficient to define the groups. Weaker forms of multiplicative separability might be $f_1(x_1,x_2)f_2(x_3,x_4)$ (calling \texttt{module(0,`[0,1]->[0,1]')}) or $f_1(x_1)f_2(x_2,x_3,x_4)$ (calling \texttt{module(0,`[0]->[0]')}).   

{\bf Generalized Symmetry} We say a function is symmetric in variables $(x_1,x_2)$ if $f(x_1,x_2,x_3,\cdots)=g(h(x_1,x_2), x_3, \cdots)$. This property is termed symmetry because the value of $f$ remains unchanged as long as  $h(x_1,x_2)$ is constant, even if $x_1$ and $x_2$ vary. For example, a function $f$ is rotational invariant in 2D if $f(x_1,x_2)=g(r)$, where $r\equiv\sqrt{x_1^2+x_2^2}$. When symmetry involves only a subset of variables, it can be considered hierarchical since $x_1$ and $x_2$ interact first through $h$ (2-Layer KAN), and then $h$ interacts with other variables via $g$ (2-Layer KAN). Suppose a four-variable function has a hierarchical form  $f(x_1,x_2,x_3,x_4)=h(f(x_1,x_2),g(x_3,x_4))$, as illustrated in Figure~\ref{fig:build_structure} (b). We can use the \texttt{module} method to create this structure by calling \texttt{module(0,`[0,1]->[0,1]->[0,1]->[0]')}, ensuring that the variable groups $(x_1, x_2)$ and $(x_3, x_4)$ do not interact in the first two layers.

\subsection{Compiling symbolic formulas to KANs}

\begin{figure}
    \centering
    \includegraphics[width=1.0\linewidth]{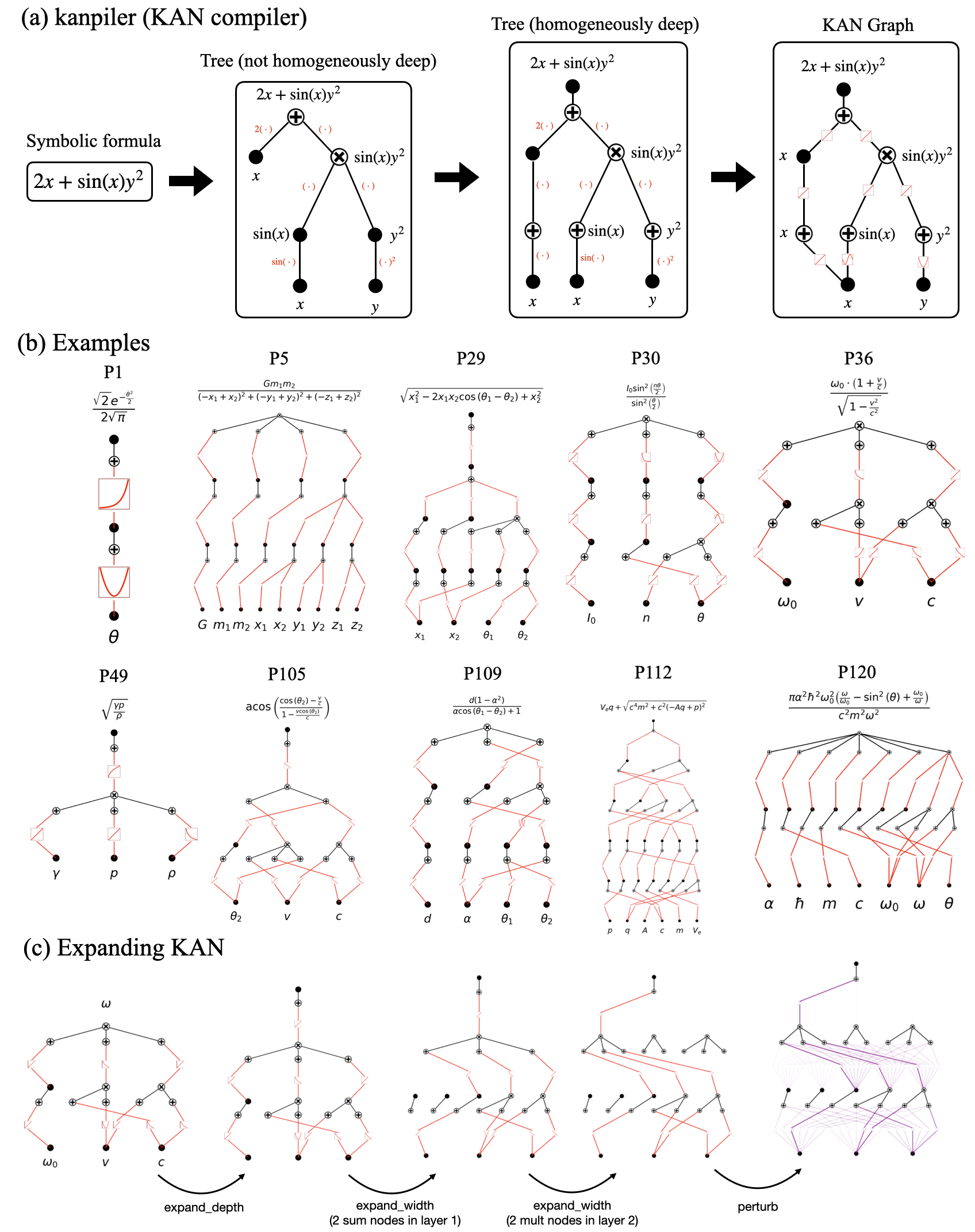}
    \caption{KAN compiler (kanpiler) converts symbolic expressions to KANs. (a) how kanpiler works: the symbolic formula is first parsed to an expression tree, which is then converted to a KAN. (b) Applying KANs to 10 equations (selected from the Feynman dataset). (c) Expand a compiled KAN to increase its expressive power.}
    \label{fig:compile}
\end{figure}

Scientists often find satisfaction in representing complex phenomena through symbolic equations. However, while these equations are concise, they may lack the expressive power needed to capture all nuances due to their specific functional forms. In contrast, neural networks are highly expressive but may inefficiently spend training time and data to learn domain knowledge already known to scientists. To leverage the strengths of both approaches, we propose a two-step procedure: (1) compile symbolic equations into KANs and (2) fine-tune these KANs using data. The first step aims to embed known domain knowledge into KANs, while the second step focuses on learning new ``physics'' from data.

{\bf kanpiler (KAN compiler)} The goal of the kanpiler is to convert a symbolic formula to a KAN. The process, illustrated in Figure~\ref{fig:compile} (a), involves three main steps: (1) The symbolic formula is parsed into a tree structure, where nodes represent expressions, and edges denote operations/functions. (2) This tree is then modified to align with the structure of a KAN graph. Modifications include moving all leaf nodes to the input layer via dummy edges, and adding dummy subnodes/nodes to match KAN architecture. These dummy edges/nodes/subnodes only perform identity transformation. (3) The variables are combined in the first layer, effectively converting the tree into a graph. For visual clarity, 1D curves are placed on edges to represent functions. We have benchmarked the kanpiler on the Feynman dataset and it successfully handles all 120 equations. Examples are shown in Figure~\ref{fig:compile} (b). The kanpiler takes input variables (as sympy symbols) and output expression (as a sympy expression), and returns a KAN model
\begin{equation}
    \texttt{model = kanpiler(input\_variables, output\_expression)}
\end{equation}
Note that the returned KAN model is in the symbolic mode, i.e., the symbolic functions are exactly encoded. If we instead use cubic splines to approximate these symbolic functions, we get MSE losses $\ell\propto N^{-8}$~\cite{liu2024kan}, where $N$ is the number of grid intervals (proportional to the number of model parameters). 

{\bf Width/depth expansion for increased expressive power} The KAN network generated by the kanpiler is compact,  without no redundant edges, which might limit its expressive power and hinder further fine-tuning. To address this, we propose \texttt{expand\_width} and \texttt{expand\_depth} methods to expand the network to become wider and deeper, as shown in Figure~\ref{fig:compile} (c). The expansion methods initially add zero activation functions, which suffer from zero gradients during training. Therefore, the  \texttt{perturb} method should be used to perturb these zero functions into non-zero values, making them trainable with non-zero gradients.

\section{KANs to Science}

Today's black box deep neural networks are powerful, but interpreting these models remains challenging. Scientists seek not only high-performing models but also the ability to extract meaningful knowledge from the models. In this section, we focus on enhancing the interpretability of KANs scientific purposes. We will explore three levels of knowledge extraction from KANs, from the most basic to the most complex: important features (Section 4.1), modular structures (Section 4.2), and symbolic formulas (Section 4.3). 

\subsection{Identifying important features from KANs}

\begin{figure}
    \centering
    \includegraphics[width=1.0\linewidth]{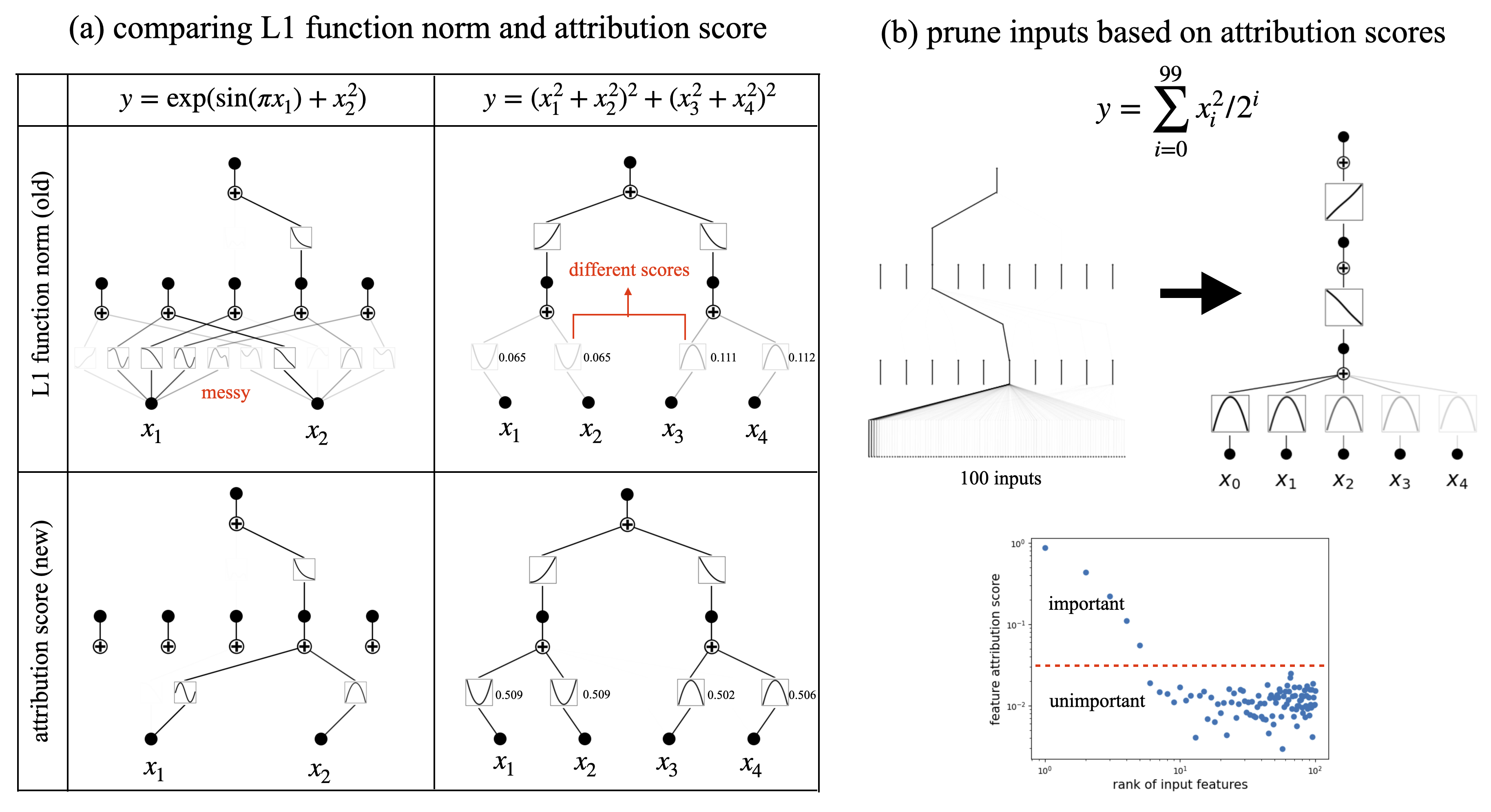}
    \caption{Identifying important features in KANs. (a) comparing the attribution score to the L1 norm used in Liu et al.~\cite{liu2024kan}. On two synthetic tasks, the attribution score brings more insights than the L1 norm. (b) Attribution scores can be computed for inputs and used for input pruning. }
    \label{fig:identify_important_features}
\end{figure}

Identifying important variables is crucial for many tasks. Given a regression model $f$ where $y\approx f(x_1,x_2,\dots,x_n)$, we aim to assign scores to the input variables to gauge their importance. Liu et al.~\cite{liu2024kan}, used the function L1 norm to indicate the importance of edges, but this metric could be problematic as it only considers local information. 

To address this, we introduce a more effective attribution score which better reflects the importance of variables than the L1 norm. For simplicity, let us assume there are multiplication nodes, so we do not need to differentiate between nodes and subnodes~\footnote{For subnodes belonging to multiplication node, the subnodes inherit their scores from the multiplication node.}. Suppose we have an $L$-layer KAN with width $[n_0, n_1,\cdots, n_L]$. We define $E_{l,i,j}$ as the standard deviation of the activations on the $(l,i,j)$ edge, and $N_{l,i}$ as the standard deviation of the activations on the $(l,i)$ node. We then define the node (attribution) score $A_{l,i}$ and the edge (attribution) score $B_{l,i,j}$. In~\cite{liu2024kan}, we simply defined  $B_{l,i,j}=E_{l,i,j}$ and $A_{l,i}=N_{l,i}$. However, this definition fails to account for the later parts of the network; even if a node or an edge has a large norm itself, it may not contribute to the output if the rest of the network is effectively a zero function. Therefore, we now compute node and edge scores iteratively from the output layer to the input layer. We set all output dimensions to have unit scores, i.e., $A_{L,i}=1$, $i=0,1,\cdots,n_L-1$~\footnote{Other choices can be made based on the perceived importance of each output dimension, though this is less critical when outputs are typically one-dimensional.}, and compute scores as follows:
\begin{equation}
        B_{l-1,i,j} = A_{l,j}\frac{E_{l,j}}{N_{l+1,j}},\quad 
        A_{l-1,i}  = \sum_{j=0}^{n_l} B_{l-1,i,j},\quad  l=L,L-1,\cdots, 1.
\end{equation}

{\bf Comparing $E_{l,i,j}$ and $B_{l,i,j}$} We find that $B_{l,i,j}$ provides a more accurate reflection of edge importance. In Figure~\ref{fig:identify_important_features}, we compare KANs trained on two equations $y={\rm exp}({\rm sin}(\pi x_1)+x_2^2)$ and $y=(x_1^2+x_2^2)^2 + (x_3^2+x_4^2)^2$ and visualize KANs with importance scores being $E$ (L1 norm) or $B$ (attribution score). For the first equation, attributions scores reveal a cleaner graph than L1 norms, as many active edges in the first layer do not contribute to the final output due to inactive subsequent edges. The attribution score accounts for this, resulting in a more meaningful graph. For the second equation $y=(x_1^2+x_2^2)^2 + (x_3^2+x_4^2)^2$, we can tell from the symbolic equation that all four variables are equally important. The attribution scores correctly reflect the equal importance of all four variables, whereas the L1 norm incorrectly suggests that $x_3$ and $x_4$ are more important than $x_1$ and $x_2$.

{\bf Pruning inputs based on attribution scores} In real datasets, input dimensionality can be large, but only a few variables may be relevant. To address this, we propose pruning away irrelevant features based on attribution scores so that we can focus on the most relevant ones. Users can apply the \texttt{prune\_input} to retain only the most relevant variables. For instance, if there are 100 input features ordered by decreasing relevance in the function  $y=\sum_{i=0}^{99}x_i^2/2^i, x_i\in [-1,1]$, and after training, only the first five features show significantly higher attribution scores, the \texttt{prune\_input} method will retain only these five features. The pruned network becomes compact and interpretable, whereas the original KAN with 100 inputs is too dense for straightforward interpretation. 

\subsection{Identifying modular structures from KANs}

Although the attribution score provides valuable insights into which edges or nodes are important, it does not reveal modular structures, i.e., how the important edges and nodes are connected. In this part, we aim to uncover modular structures from trained KANs and MLPs by examining two types of modularity: anatomical modularity and functional modularity. 

\subsubsection{Anatomical modularity}

\begin{figure}
    \centering
    \includegraphics[width=1.0\linewidth]{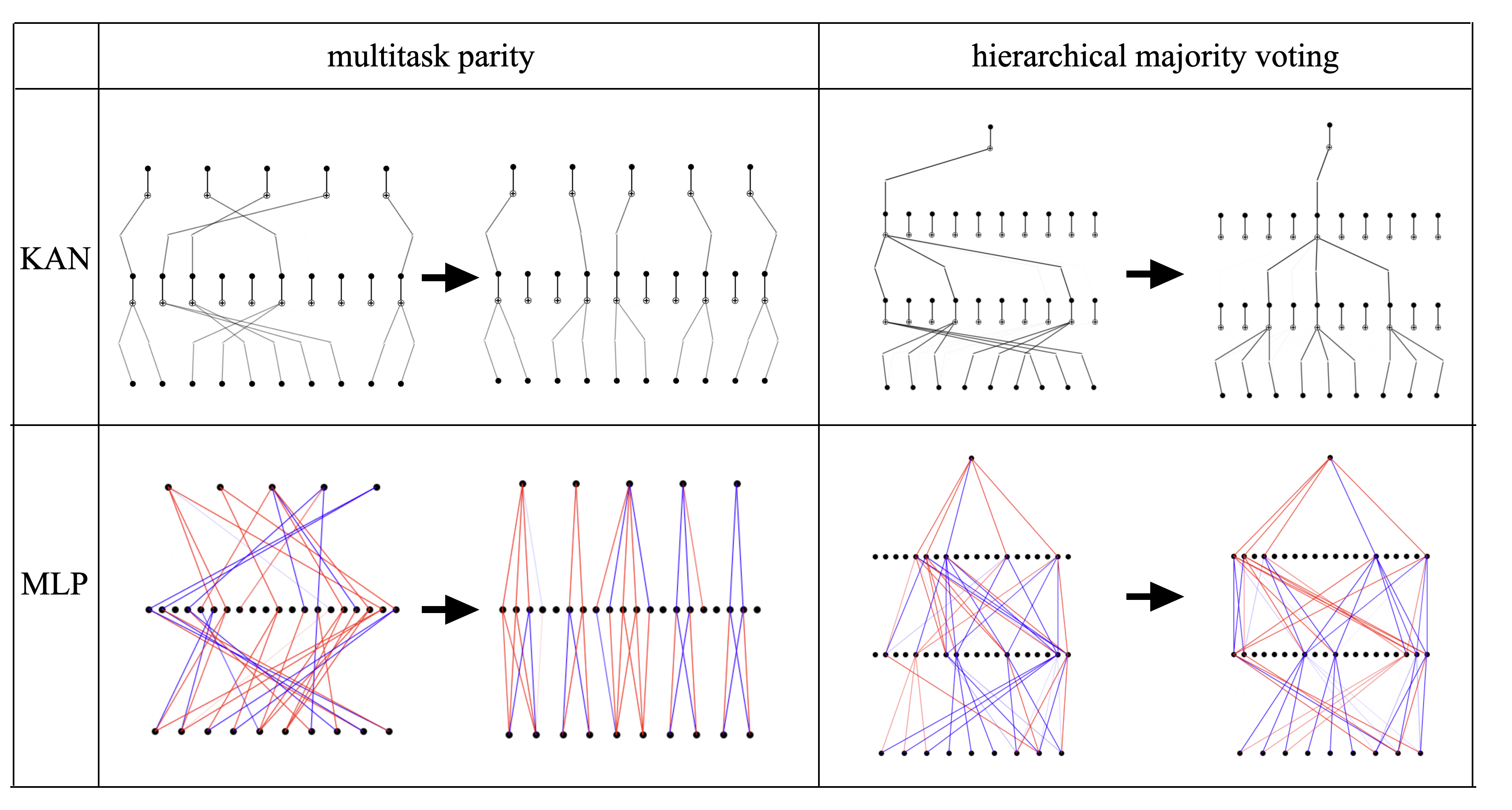}
    \caption{Inducing anatomical modularity in neural networks through neuron swapping. The approach involves assigning spatial coordinates to neurons and permuting them to minimize the overall connection cost. For two tasks (left: multitask parity, right: hierarchical majority voting), neuron swapping works for KANs (top) in both cases and works for MLPs (bottom) for multitask parity.}
    \label{fig:anatomical-modularity}
\end{figure}

Anatomical modularity refers to the tendency for neurons placed close to each other spatially to have stronger connections than those further apart. Although artificial neural networks lack physical spatial coordinates, introducing the concept of physical space has been shown to enhance interpretability~\cite{liu2023seeing, liu2023growing}. We adopt the neuron swapping method from~\cite{liu2023seeing, liu2023growing}, which shortens connections while preserving the network's functionality. We call the method \texttt{auto\_swap}. The anatomical modular structure revealed through neuron swapping facilitates easy identification of modules, even visually, for two tasks shown Figure~\ref{fig:anatomical-modularity}: (1) multitask sparse parity; and (2) hierarchical majority voting. For multitask sparse parity, we have 10 input bits  $x_i\in\{0,1\}, i=1,2,\cdots, 10$, and output $y_j=x_{2j-1}\oplus x_{2j}, j=1,\cdots,5$, where $\oplus$ denotes modulo 2 addition. The task exhibits modularity because each output depends only on a subset of inputs. \texttt{auto\_swap} successfully identifies modules for both KANs and MLPs, with the KAN discovering simpler modules. For hierarchical majority voting, with 9 input bits $x_i\in\{0,1\}, i=1,\cdots,9$, and the output $y={\rm maj}({\rm maj}(x_1,x_2,x_3),{\rm maj}(x_4,x_5,x_6),{\rm maj}(x_7,x_8,x_9))$, where ${\rm maj}$ stands for majority voting (output 1 if two or three inputs are 1, otherwise 0). The KAN reveals the modular structure even before \texttt{auto\_swap}, and the diagram becomes more organized after \texttt{auto\_swap}. The MLP shows some modular structure from the pattern of the first layer weights, indicating interactions among variables, but the global modular structure remains unclear regardless of \texttt{auto\_swap}.

\subsubsection{Functional modularity} 

\begin{figure}
    \centering
    \includegraphics[width=1.0\linewidth]{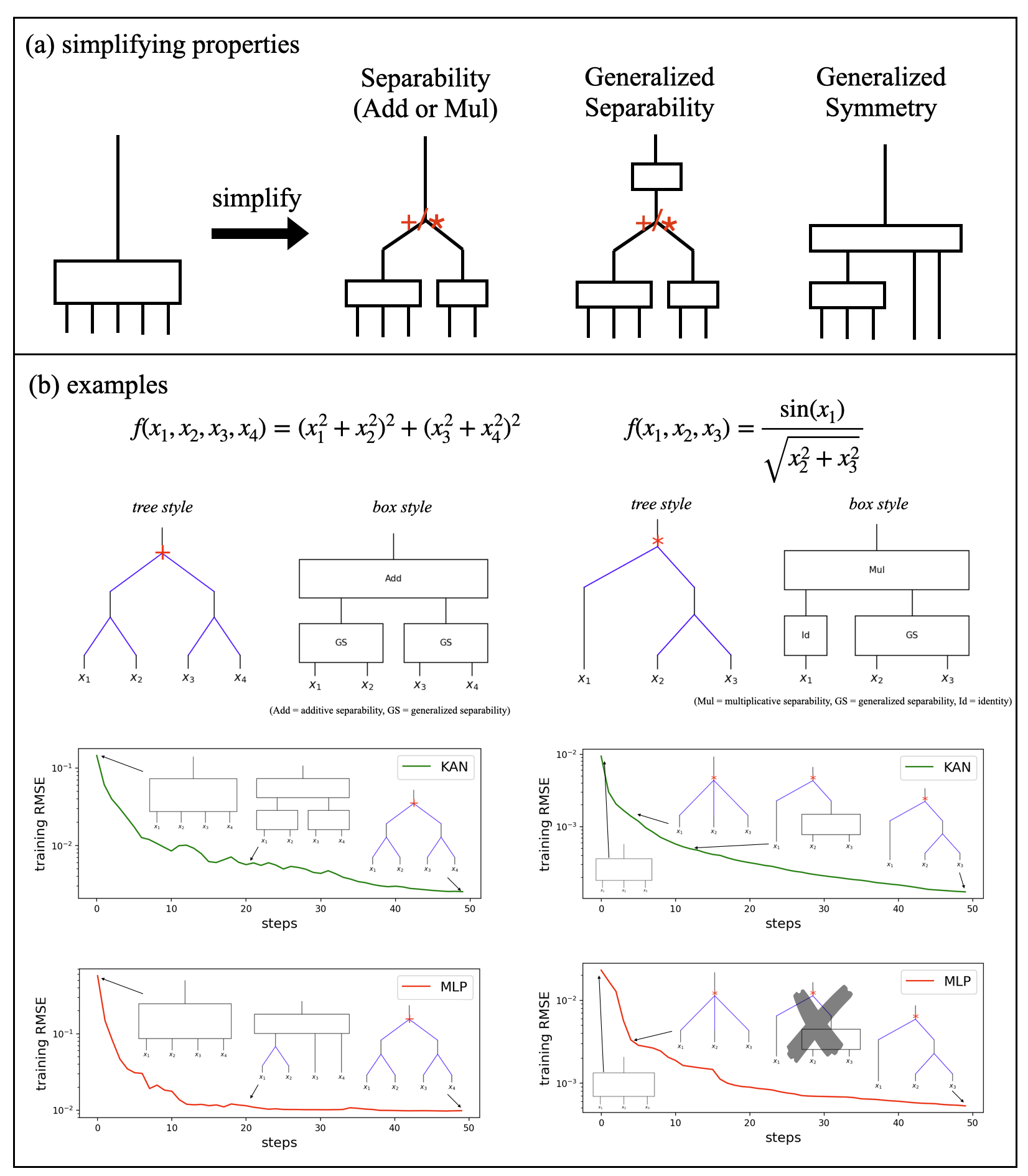}
    \caption{Detecting functional modularity in KANs. (a) We study three types of functional modularity: separability (additive or multiplicative), general separability, and symmetry. (b) Applying these tests recursively converts a function into a tree. Here the function can be symbolic functions (top), KANs (middle) or MLPs (bottom). Both KANs and MLPs produce correct tree graphs at the end of training but show different training dynamics.}
    \label{fig:functional_modularity}
\end{figure}

Functional modularity pertains to the overall function represented by the neural network. Given an Oracle network where internal details such as weights and hidden layer activations are inaccessible (too complicated to analyze), we can still gather information about functional modularity through forward and backward passes at the inputs and outputs. We define three types of functional modularity (see Figure~\ref{fig:functional_modularity} (a)), based largely on ~\cite{udrescu2020ai}.

{\bf Separability}: A function $f$ is additively separable if 
\begin{equation}
    f(x_1,x_2,\cdots x_n) = g(x_1,\dots, x_{k}) + h(x_{k+1},\dots, x_n).
\end{equation}
Note that $\frac{\partial^2 f}{\partial x_i\partial x_j}=0$ when $1\leq i\leq k, k+1\leq j\leq n$. To detect the separability, we can compute the Hessian matrix $\mat{H}\equiv\nabla^T\nabla f\ (\mat{H}_{ij}=\frac{\partial^2 f}{\partial x_i\partial x_j})$ and check for block structure. If $\mat{H}_{ij}=0$ for all $1\leq i\leq k$ and $k+1\leq j\leq n$, then we know $f$ is additively separable. For multiplicative separability, we can convert it to additive separability by taking the logarithm:
\begin{equation}
\begin{aligned}
    & f(x_1,x_2,\cdots x_n) = g(x_1,\dots, x_{k}) \times  h(x_{k+1},\dots, x_n) \\
    & {\rm log}\left|f(x_1,x_2,\cdots, x_n)\right| = {\rm log}\left|g(x_1,\dots, x_{k})\right| + {\rm log}\left|h(x_{k+1},\dots, x_n)\right|
\end{aligned}
\end{equation}
To detect multiplicative separability, we define $\mat{H}_{ij}\equiv\frac{\partial^2{\rm log}|f|}{\partial x_i\partial x_j}$, and check for block structure. Users can call \texttt{test\_separability} to test general separability.

{\bf Generalized separability}: A function $f$ has generalized separability if 
\begin{equation}
    f(x_1,x_2,\cdots x_n) = F(g(x_1,\dots, x_{k}) + h(x_{k+1},\dots, x_n)).
\end{equation}
To detect generalized separability, we compute
\begin{equation}
\begin{aligned}
    & \frac{\partial f}{\partial x_i} = \frac{\partial F}{\partial g}\frac{\partial g}{\partial x_i}\ (1\leq i\leq k),\  \frac{\partial f}{\partial x_j} = \frac{\partial F}{\partial h}\frac{\partial h}{\partial x_i}\ (k+1\leq j\leq n) \\
    & \frac{\partial f/\partial x_i}{\partial f/\partial x_j} = \frac{\partial F/\partial g}{\partial F/\partial h}\frac{\partial g/\partial x_i}{\partial h/\partial x_j} = \frac{\partial g/\partial x_i}{\partial h/\partial x_j} = g_{x_i}(x_1,x_2,\cdots x_k)\times \frac{1}{h_{x_j}(x_{k+1},\cdots, x_n)}.
\end{aligned} 
\end{equation}
where we have used $\frac{\partial F}{\partial g}=\frac{\partial F}{\partial h}$. Note that $\frac{\partial f/\partial x_i}{\partial f/\partial x_j}$ is multiplicatively separable, it can be detected by the separability test proposed above. Users can call \texttt{test\_general\_separability} to check for additive or multiplicative separability.  

{\bf Generalized Symmetry}: 
A function has generalized symmetry (in the first $k$ variables) if
\begin{equation}
    f(x_1,x_2,\cdots,x_n) = g(h(x_1,\cdots, x_k), x_{k+1},\cdots, x_n).
\end{equation}
We denote $\mat{y}=(x_1,\cdots, x_k)$ and $\mat{z}=(x_{k+1},\cdots, x_n)$. This property is called generalized symmetry because $f$ retains the same value as long as $h$ is held constant, regardless of individual values of $x_1, \cdots, x_k$. We compute the gradient of $f$ with respect to $\mat{y}$: $\nabla_{\mat{y}}f = \frac{\partial g}{\partial h} \nabla_{\mat{y}}h$. Since $\frac{\partial g}{\partial h}$ is a scalar function, it does not change the direction of $\nabla_{\mat{y}}h$. Thus, the direction of  $\widehat{\nabla_{\mat{y}}f} \equiv \frac{\nabla_{\mat{y}}f}{|\nabla_{\mat{y}}f|}$ is independent of $\mat{z}$, i.e., 
\begin{equation}
\nabla_\mat{z}(\widehat{\nabla_{\mat{y}}f}) = 0,
\end{equation}
which is the condition for symmetry. Users can call the \texttt{test\_symmetry} method to check for symmetries.

{\bf Tree converter} The three types of functional modularity form a hierarchy: symmetry is the most general, general separability is intermediate, and separability is the most specific. Mathematically,
\begin{equation}
    {\rm Separability} \subset {\rm Generalized\  Separability} \subset {\rm Generalized\ Symmetry}
\end{equation}
To obtain the maximal hierarchy of modular structures, we apply generalized symmetry detection recursively, forming groups as small as $k=2$ variables and extending to all $k=n$ variables. For example, let us consider an 8-variable function
\begin{equation}
    f(x_1, \cdots, x_8) = ((x_1^2+x_2^2)^2 + (x_3^2+x_4^2)^2)^2 + ((x_5^2+x_6^2)^2 + (x_7^2+x_8^2)^2)^2,
\end{equation}
which has four $k=2$ generalized symmetries, involving groups $(x_1,x_2)$, $(x_3,x_4)$, $(x_5, x_6)$, $(x_7, x_8)$; two $k=2$ generalized symmetries, involving groups $(x_1, x_2, x_3, x_4)$ and $(x_5,x_6,x_7,x_8)$. As such, each $k=4$ group contains two $k=2$ groups, demonstrating a hierarchy. For each generalized symmetry, we can also test if the generalized symmetry is further generalized separable or separable. Users can use the method \texttt{plot\_tree} to obtain the tree graph for a function (the function could be any Python expressions, neural networks, etc.). For a neural network \texttt{model}, users can simply call \texttt{model.tree()}.  The tree plot can have the style `tree' (by default) or `box'. 

{\bf Examples} Figure~\ref{fig:functional_modularity} (b) provides two examples. When the exact symbolic functions are input to \texttt{plot\_tree}, the ground truth tree graphs are obtained. We are particularly interested in whether the tree converter works for neural networks. For these simple cases,  both KANs and MLPs can find the correct graph if sufficiently trained. Figure~\ref{fig:functional_modularity} (b) (bottom) shows the evolution of the tree graphs during KAN and MLP training. It is particularly interesting to see how neural networks gradually learn the correct modular structure. In the first case $f(x_1,x_2,x_3,x_4) = (x_1^2+x_2^2)^2+(x_3^2+x_4^2)^2$, both KAN and MLP gradually pick up more inductive biases (their intermediate states are different) until they reach the correct structure. In the second case,  $f(x_1,x_2,x_3)={\rm sin}(x_1)/\sqrt{x_2^2+x_3^2}$, both the models initially detect multiplicative separability for all three variables, showing even higher symmetry than the correct structure. After training progresses, both models ``realize'' that: in order to better fit data (loss becomes lower), such high symmetry structure can no longer be met and should be relaxed to a less stringent structure. An additional observation is that KAN has an intermediate structure not found in the MLP. There are two caveats we would like to mention: (1) results can be seed and/or threshold-dependent. (2) all tests rely on second-order derivatives, which may not be robust due to the model being trained only on zero-order information. Adversarial constructions such as $f_\epsilon(x)=f(x) + \epsilon {\rm sin}(\frac{x}{\epsilon})$ could lead to issues, because although $|f_\epsilon(x)-f(x)|\to 0$ as $\epsilon\to 0$, $|f''_\epsilon(x)-f''(x)|\to \infty$ as $\epsilon\to 0$. Although such extreme cases are unlikely in practice, smoothness is necessary to ensure the success of our methods. 

\subsection{Identifying symbolic formulas from KANs}

\begin{figure}
    \centering
    \includegraphics[width=1.0\linewidth]{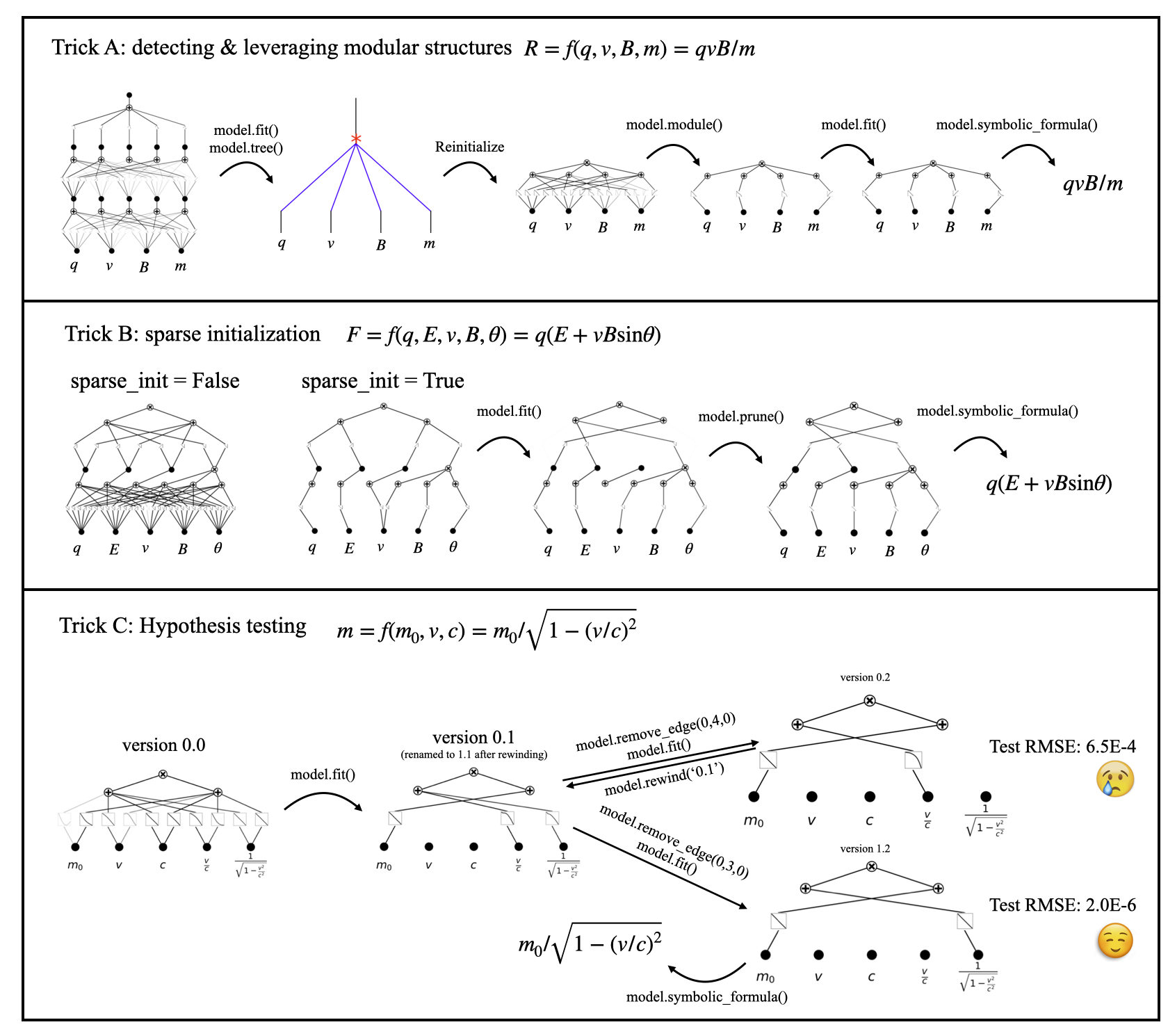}
    \caption{Three tricks to facilitate symbolic regression. Trick A (top row): detecting and leveraging modular structures. Trick B (middle row): sparse connection initialization. Trick C (bottom row): Hypothesis testing.}
    \label{fig:symbolic_regression}
\end{figure}

Symbolic formulas are the most informative, as they clearly reveal both important features and modular structures once they are known. In Liu et al.~\cite{liu2024kan}, the authors showed a bunch of examples from which they can extract symbolic formulas, with some prior knowledge when needed. With the new tools proposed above (feature importance, modular structures, and symbolic formulas), users can leverage these new tools to easily interact and collaborate with KANs, making symbolic regression easier. We present three tricks below, illustrated in Figure~\ref{fig:symbolic_regression}. 

{\bf Trick A: discover and leverage modular structures}
We can first train a general network and probe its modularity. Once the modular structure is identified, we initialize a new model with this modular structure as inductive biases. For instance, consider the function $f(q,v,B,m)=qvB/m$. We first initialize a large KAN (presumably expressive enough) to fit the dataset to a reasonable accuracy. After training, the tree graph is extracted (ref Sec 4.2) from the trained KAN, which shows multiplicative separability. Then we can build the modular structure into a second KAN (ref Sec 3.2), train it, and then symbolify all 1D functions to derive the formula. 

{\bf Trick B: Sparse initialization}
Symbolic formulas typically correspond to KANs with sparse connections (see Figure~\ref{fig:compile} (b)), so initializing KANs sparsely aligns them better with the inductive biases of symbolic formulas. Otherwise, densely initialized KANs require careful regularization to promote sparsity. Sparse initialization can be achieved by passing the argument ``\texttt{sparse\_init=True}'' to the KAN initializer. For example, for the function $f(q,E,v,B,\theta)=q(E+vB{\rm sin}\theta)$, a sparsely initialized KAN closely resembles the final trained KAN, requiring only minor adjustments in training. In contrast, a dense initialization would involve extensive training to remove unnecessary edges.

{\bf Trick C: Hypothesis Testing}
When faced with multiple reasonable hypotheses, we can try all of them (branching into ``parallel universes'') to test which hypothesis is the most accurate and/or simplest. To facilitate hypothesis testing, we build a checkpoint system that automatically saves model versions whenever changes (e.g., training, pruning) are made. For example, consider the function $f(m_0,v,c)=m_0/\sqrt{1-(v/c)^2}$. We start from a randomly initialized KAN, which has version 0.0. After training, it evolves to version 0.1, where it activates on both $\beta=v/c$ and $\gamma=1/\sqrt{1-(v/c)^2}$. Hypothesize that only $\beta$ or $\gamma$ might be needed. We first set the edge on $\gamma$ to zero, and train the model, obtaining a $6.5\times 10^{-4}$ test RMSE (version 0.2). To test the alternative hypothesis, we want to revert back to the branching point (version 0.1) -- we call \texttt{model.rewind(`0.1')} which rewinds the model back to version 0.1. To indicate that \texttt{rewind} is called, version 0.1 is renamed to version 1.1. Now we set the edge on $\beta$ to be zero, train the model, obtaining a $2.0\times 10^{-6}$ test RMSE (the version becomes 1.2). Comparing versions 0.2 and 1.2 indicates that the second hypothesis is better due to the lower loss given the same complexity (both hypotheses have two non-zero edges).

\section{Applications}

The previous sections primarily focused on regression problems for pedagogical purposes. In this section, we apply KANs to discover physical concepts, such as conserved quantities, Lagrangians, hidden symmetries, and constitutive laws. These examples illustrate how the tools proposed in this paper can be effectively integrated into real-life scientific research to tackle these complex tasks. 

\subsection{Discovering conserved quantities}
\begin{figure}[htbp]
    \centering
    \includegraphics[width=1.0\linewidth]{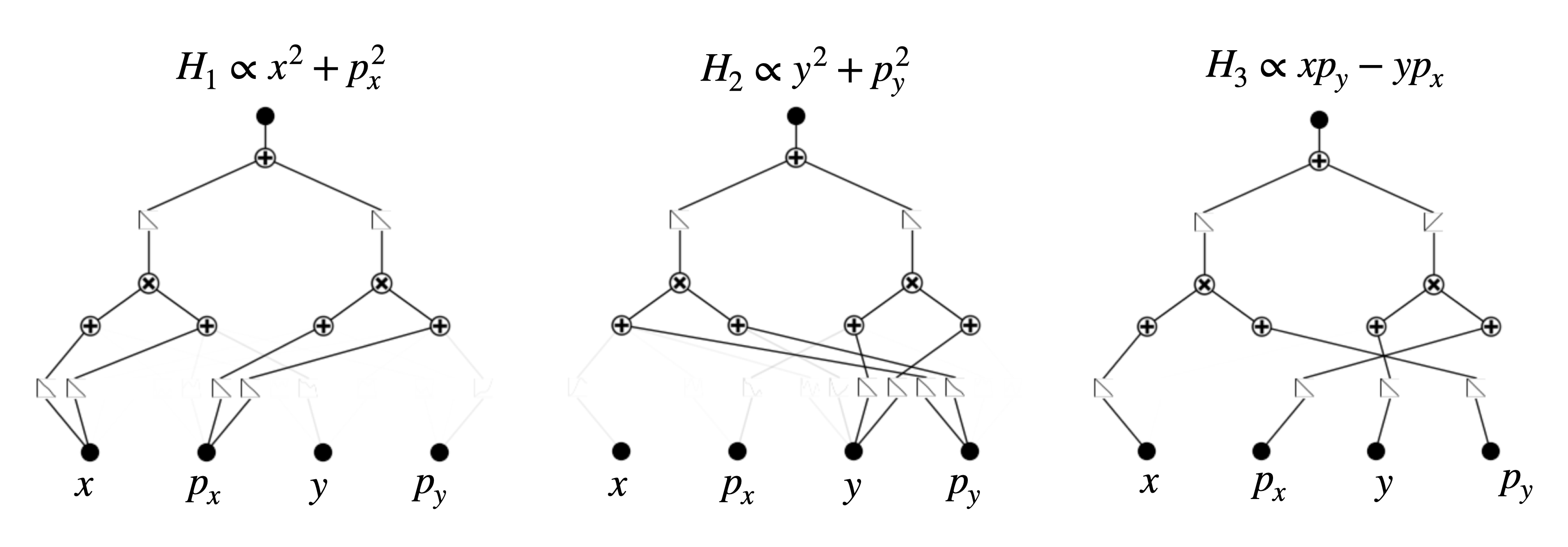}
    \caption{Using KANs to discover conserved quantities for the 2D harmonic oscillator.}
    \label{fig:poincare}
\end{figure}
Conserved quantities are physical quantities that remain constant over time. For example, a free-falling ball converts its gravitational potential energy into kinetic energy, while the total energy (the sum of both forms of energy) remains constant (assuming negligible air resistance). Conserved quantities are crucial because they often correspond to symmetries in physical systems and can simplify calculations by reducing the dimensionality of the system. Traditionally, deriving conserved quantities with paper and pencil can be time-consuming and demands extensive domain knowledge. Recently, machine learning techniques have been explored to discover conserved quantities~\cite{aipoincare, aipoincare2, sid, lu2023discovering, ha2021discovering, sebastian-cl}. 

We follow the approach Liu et al.~\cite{aipoincare2}, which derived a differential equation that conserved quantities must satisfy, thus transforming the problem of finding conserved quantities into differential equation solving. They used multi-layer perceptrons (MLPs) to parameterize conserved quantities. We basically follow their procedure but replace MLPs with KANs. To be specific, they consider a dynamical system with the state variable $\mat{z}\in\mathbb{R}^d$ governed by the equation $\frac{d\mat{z}}{dt}=\mat{f}(\mat{z})$. The necessary and sufficient condition for a function $H(\mat{z})$ to be a conserved quantity is that $\mat{f}(\mat{z})\cdot\nabla H(\mat{z})=0$ for all $\mat{z}$. For example, in a 1D harmonic oscillator, the phase space is characterized by position and momentum, $\mat{z}=(x,p)$, and the evolution equation is $d(x,p)/dt=(p,-x)$. The energy $H=\frac{1}{2}(x^2+p^2)$ is a conserved quantity because $\mat{f}(\mat{z})\cdot\nabla H(\mat{z})=(p,-x)\cdot(x,p)=0$. We parameterize $H$ using a KAN, and train it with the loss function $\ell=\sum_{i=1}^N\left|\mat{f}(\mat{z}^{(i)})\cdot\widehat{\nabla}H(\mat{z}^{(i)})\right|^2$ where $\widehat{\nabla}$ is the normalized gradient, and $\mat{z}^{(i)}$ are the $i^{\rm th}$ data point uniformly drawn from the hypercube $[-1,1]^d$. 

We choose the 2D harmonic oscillator to test KANs, characterized by $(x, y, p_x, p_y)$. It has three conserved quantities: (1) energy along $x$ direction: $H_1=\frac{1}{2}(x^2+p_x^2)$; (2) energy along $y$ direction: $H_2=\frac{1}{2}(y^2+p_y^2)$; (3) angular momentum $H_3 = xp_y - yp_x$. We train $[4,[0,2],1]$ KANs with three different random seeds, as shown in Figure~\ref{fig:poincare}, which correspond to $H_1$, $H_2$ and $H_3$ respectively.

\subsection{Discovering Lagrangians}

\begin{figure}[htbp]
    \centering
    \includegraphics[width=1.0\linewidth]{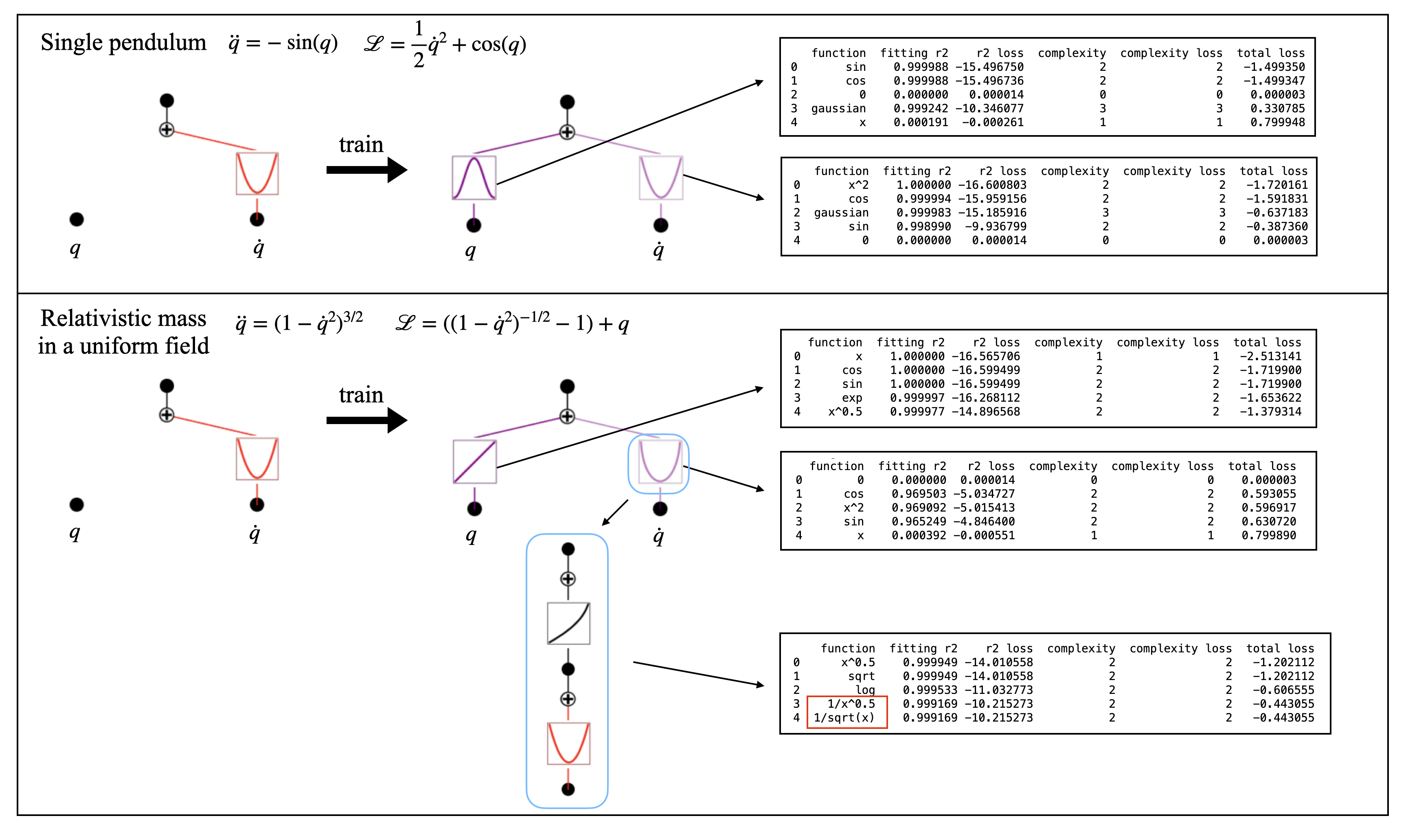}
    \caption{Use KANs to learn Lagrangians for the single pendulum (top) and a relativistic mass in a uniform field (bottom).}
    \label{fig:lnn}
\end{figure}

In physics, Lagrangian mechanics is a formulation of classical mechanics based on the principle of stationary action. It describes a mechanical system using phase space and a smooth function $\mathcal{L}$ known as the Lagrangian. For many systems, $\mathcal{L}=T-V$, where $T$ and $V$ represent the kinetic and potential energy of the system, respectively. The phase space is typically described by $(\mat{q}, \dot{\mat{q}})$, where $\mat{q}$ and $\dot{\mat{q}}$ denotes coordinates and velocities, respectively. The equation of motion can be derived from the Lagrangian via the Euler-Lagrange equation: $\frac{d}{dt}(\frac{\partial\mathcal{L}}{\partial\dot{\mat{q}}})=\frac{\partial\mathcal{L}}{\partial \mat{q}}$, or equivalently 
\begin{equation}\label{eq:qdd}
\ddot{\mat{q}}=(\nabla_{\dot{\mat{q}}}\nabla^T_{\dot{\mat{q}}}\mathcal{L})^{-1}[\nabla_{\mat{q}}\mathcal{L}-(\nabla_{\mat{q}}\nabla^T_{\dot{\mat{q}}}\dot{\mat{q})}]
\end{equation}
Given the fundamental role of the Lagrangian, an interesting question is whether we can infer the Lagrangian from data. Following~\cite{cranmer2020lagrangian}, we train a Lagrangian neural network to predict $\ddot{\mat{q}}$ from $(\mat{q}, \dot{\mat{q}})$. An LNN uses an MLP to parameterize $\mathcal{L}(\mat{q},\dot{\mat{q}})$, and computes the Eq.~(\ref{eq:qdd}) to predict instant accelerations $\ddot{\mat{q}}$. However, LNNs face two main challenges: (1) The training of LNNs can be unstable due to the second-order derivatives and matrix inversion in Eq.~(\ref{eq:qdd}). (2) LNNs lack interpretability because MLPs themselves are not easily interpretable. We address these issues using KANs.

To tackle the first challenge, we note that the matrix inversion of the Hessian $(\nabla_{\dot{\mat{q}}}\nabla^T_{\dot{\mat{q}}}\mathcal{L})^{-1}$ becomes problematic when the Hessian has eigenvalues close to zero. To mitigate this, we initialize $(\nabla_{\dot{\mat{q}}}\nabla^T_{\dot{\mat{q}}}\mathcal{L})$ as a positive definite matrix (or a positive number in 1D). Since  $(\nabla_{\dot{\mat{q}}}\nabla^T_{\dot{\mat{q}}}\mathcal{L})$ is the mass $m$ in classical mechanics and kinetic energy is usually $T = \frac{1}{2}m\dot{\mat{q}}^2$, encoding this prior knowledge into KANs is more straightforward than into MLPs (using the kanpiler introduced in Section 3.3). The kanpiler can convert the symbolic formula $T$ into a KAN (as shown in Figure~\ref{fig:lnn}). We use this converted KAN for initialization and continue training, resulting in much greater stability compared to random initialization. After training, symbolic regression can be applied to each edge to extract out symbolic formulas, addressing the second challenge.

We show two 1D examples in Figure~\ref{fig:lnn}, a single pendulum and a relativistic mass in a uniform field. The compiled KANs are displayed on the left, with edges on $\dot{q}$ displaying quadratic functions and edges on $q$ as zero functions.

{\bf Single pendulum} The $\dot{q}$ part remains a quadratic function $T(\dot{q})=\frac{1}{2}\dot{q}^2$ while the $q$ part learns to be a cosine function, as $V(q)=1-{\rm cos}(q)$. In Figure~\ref{fig:lnn} top, the results from \texttt{suggest\_symbolic} display the top five functions that best match the splines, considering both fitness and simplicity. As expected, the cosine and the quadratic function appear at the top of the lists. 

{\bf Relativistic mass in a uniform field} After training, the kinetic energy part deviates from $T=\frac{1}{2}\dot{q}^2$ because, for a relativistic particle, $T_r=(1-\dot{q}^2)^{-1/2}-1$. In Figure~\ref{fig:lnn} (bottom), symbolic regression successfully finds $V(q)=q$, but fails to identify $T_r$ due to its compositional nature, as our symbolic regression only searches for simple functions. By assuming the first function composition is quadratic, we create another $[1,1,1]$ KAN to fit $T_r$ and set the first function to be the quadratic function using \texttt{fix\_symbolic}, and train only the second learnable function. After training, we see that the ground truth $x^{-1/2}$ appears among the top five candidates. However, $x^{1/2}$ fits the spline slightly better, as indicated by a higher R-squared value. This suggests that symbolic regression is sensitive to noise (due to imperfect learning) and prior knowledge is crucial for correct judgment. For instance, knowing that kinetic energy should diverge as velocity approaches the speed of light helps confirm $x^{-1/2}$ as the correct term, since $x^{1/2}$ does not exhibit the expected divergence.

\subsection{Discovering hidden symmetry}

\begin{figure}[htbp]
    \centering
    \includegraphics[width=1.0\linewidth]{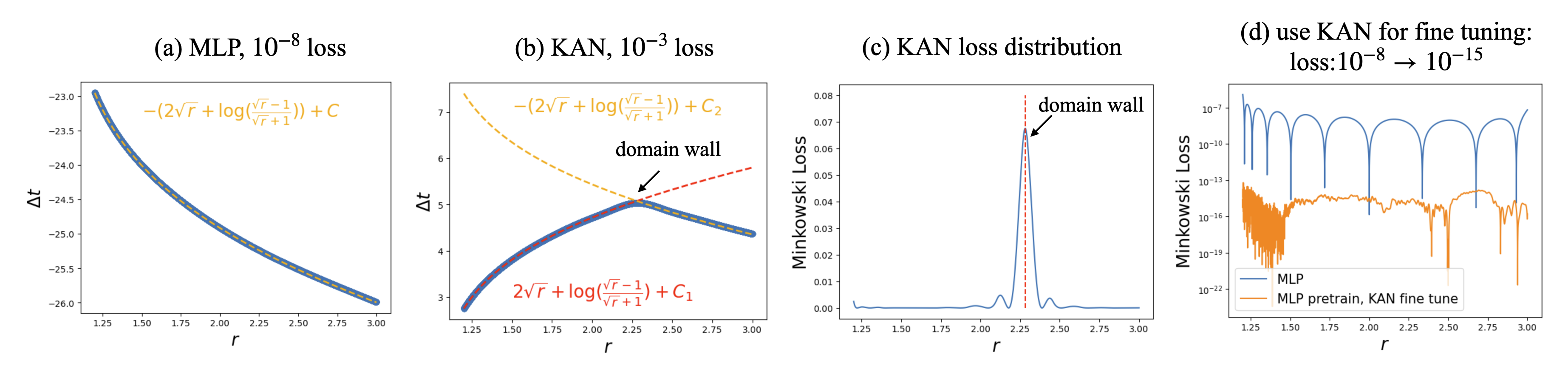}
    \caption{Rediscovering the hidden symmetry of the Schwarzschild black hole with MLPs and KANs. (a) $\Delta t(r)$ learned by the MLP is a globally smooth solution; (b) $\Delta t(r)$ learned by the KAN is a domain-wall solution; (c) The KAN shows a loss spike at the domain wall; (d) A KAN can be used to fine-tune the MLP solution close to machine precision.}
    \label{fig:hidden-symmetry}
\end{figure}

Philip Anderson famously argued that ``it is only slightly overstating that case to say that physics is the study of symmetry'', emphasizing how the discovery of symmetries has been invaluable for both deepening our understanding and solving problems more efficiently. 

However, symmetries are sometimes not manifest but \textit{hidden}, only revealed by applying some coordinate transformation. For example, after Schwarzschild discovered his eponymous black hole metric, it took 17 years for 
Painlev\'e, 
Gullstrand 
and Lema\^itre 
to uncover its hidden translational symmetry.
They demonstrated that the spatial sections could be made translationally invariant with a clever coordinate transformation, thereby deepening our understanding of black holes \cite{MisnerThorneWheelerBook}. Liu \& Tegmark~\cite{liu2022hiddensymmetry} showed that the Gullstrand-Painlev\'e transformation can be discovered by training an MLP in minutes. However, they did not get extremely high precision (i.e., machine precision) for the solution. We attempt to revisit this problem using KANs.

Suppose there is a Schwarzschild black hole in spacetime $(t,x,y,z)$ with mass $2M=1$,  centered at $x=y=z=0$ with a radius $r_s = 2M =1$. The Schwarzschild metric describes how space and time distorts around it:
\begin{equation}
    \mat{g}_{\mu\nu}=\begin{pmatrix}
    1-\frac{2M}{r} & 0 & 0 & 0 \\
    0 & -1-\frac{2Mx^2}{(r-2M)r^2} & -\frac{2Mxy}{(r-2M)r^2} & -\frac{2Mxz}{(r-2M)r^2} \\
    0 & -\frac{2Mxy}{(r-2M)r^2} & -1-\frac{2My^2}{(r-2M)r^2} & -\frac{2Myz}{(r-2M)r^2} \\
    0 & -\frac{2Mxz}{(r-2M)r^2} & -\frac{2Myz}{(r-2M)r^2} & -1-\frac{2Mz^2}{(r-2M)r^2}.
    \end{pmatrix}
\end{equation}
Applying the Gullstrand-Painlev\'e transformation $t' = t + 2M(2u+{\rm ln}(\frac{u-1}{u+1})), u\equiv\sqrt{\frac{r}{2M}}$, $x'=x$, $y'=y$, $z'=z$, the metric in the new coordinates becomes:
\begin{equation}
    \mat{g}'_{\mu\nu}=\begin{pmatrix}
    	1-\frac{2M}{r} & -\sqrt{\frac{2M}{r}}\frac{x}{r} & -\sqrt{\frac{2M}{r}}\frac{y}{r} & -\sqrt{\frac{2M}{r}}\frac{z}{r} \\
    	-\sqrt{\frac{2M}{r}}\frac{x}{r} & -1 & 0 & 0 \\
    	-\sqrt{\frac{2M}{r}}\frac{y}{r} & 0 & -1 & 0 \\
    	-\sqrt{\frac{2M}{r}}\frac{z}{r} & 0 & 0 & -1
    	\end{pmatrix},
\end{equation}
which exhibits translation invariance in the spatial section (the lower right $3\times 3$ block is the Euclidean metric). Liu \& Tegmark~\cite{liu2022hiddensymmetry} used an MLP to learn the mapping from $(t, x, y, z)$ to $(t', x', y', z')$. Defining the Jacobian matrix $\mat{J}\equiv \frac{\partial (t', x' ,y' ,z')}{\partial (t,x,y,z)}$, $\mat{g}$ is tranformed to $\mat{g}'=\mat{J}^{-T}\mat{g}\mat{J}^{-1}$. We take the bottom right $3\times 3$ block of $\mat{g}'$ and take its difference to the Euclidean metric to obtain the MSE loss. The loss is minimized by doing gradient descents on the MLP. To make things simple, they assume knowing $x'=x, y'=y, z'=z$, and only use an MLP (1 input and 1 output) to predict the temporal difference $\Delta t(r) = t'-t = 2M(2u+{\rm ln}(\frac{u-1}{u+1})), u\equiv\sqrt{\frac{r}{2M}}$ from the radius $r$. 

{\bf MLP and KAN find different solutions} We trained both an MLP and a KAN to minimize this loss function, with results shown in Figure~\ref{fig:hidden-symmetry}. Since the task has 1 input dimension and 1 output dimension, the KAN effectively reduces to a spline. We originally expected KANs to outperform MLPs, because splines are known to be superior in low-dimensional settings~\cite{michaud2023precision}. However, while MLP can achieve $10^{-8}$ loss, the KAN gets stuck at $10^{-3}$ loss despite grid refinements. It turned out that KAN and MLP learned two different solutions: while the MLP found a globally smooth solution (Figure~\ref{fig:hidden-symmetry} (a)), the KAN learned a domain-wall solution (Figure~\ref{fig:hidden-symmetry} (b)). The domain wall solution has a singular point that separates the whole curve into two segments. The left segment learns $\Delta t(r)$ correctly, while the right segment learns $-\Delta t(r)$, which is also a valid solution but differs from the left segment by a minus sign. There is a loss spike appearing at the singular point (Figure~\ref{fig:hidden-symmetry} (c)). One might consider this as a feature of KANs because domain wall solutions are prevalent in nature. However, if one considers this a flaw, KANs can still obtain globally smooth solutions by adding regularizations (to reduce spline oscillations) or experimenting with different random seeds (roughly 1 out of 3 random seeds finds a global smooth solution).

{\bf KANs can achieve extreme precision} Although the MLP finds the globally smooth solution and achieves $10^{-8}$ loss, the loss is still far from machine precision. We found that neither longer training nor increasing the MLP's size significantly reduced the loss. Therefore, we turned to KANs, which, as splines in 1D, can achieve arbitrary accuracy by refining the grid (given infinite data). We first used the MLP as a teacher, generating supervised pairs $(x,y)$ to train the KAN to fit the supervised data. This way, the KAN is initialized to a globally smooth solution. We then iteratively refined the KAN by increasing the number of grid intervals to 1000. In the end, the fine-tuned KANs achieve a loss of $10^{-15}$, close to machine precision (Figure~\ref{fig:hidden-symmetry} (d)). 

\subsection{Learning constitutive laws}

\begin{figure}
    \centering
    \includegraphics[width=1.0\linewidth]{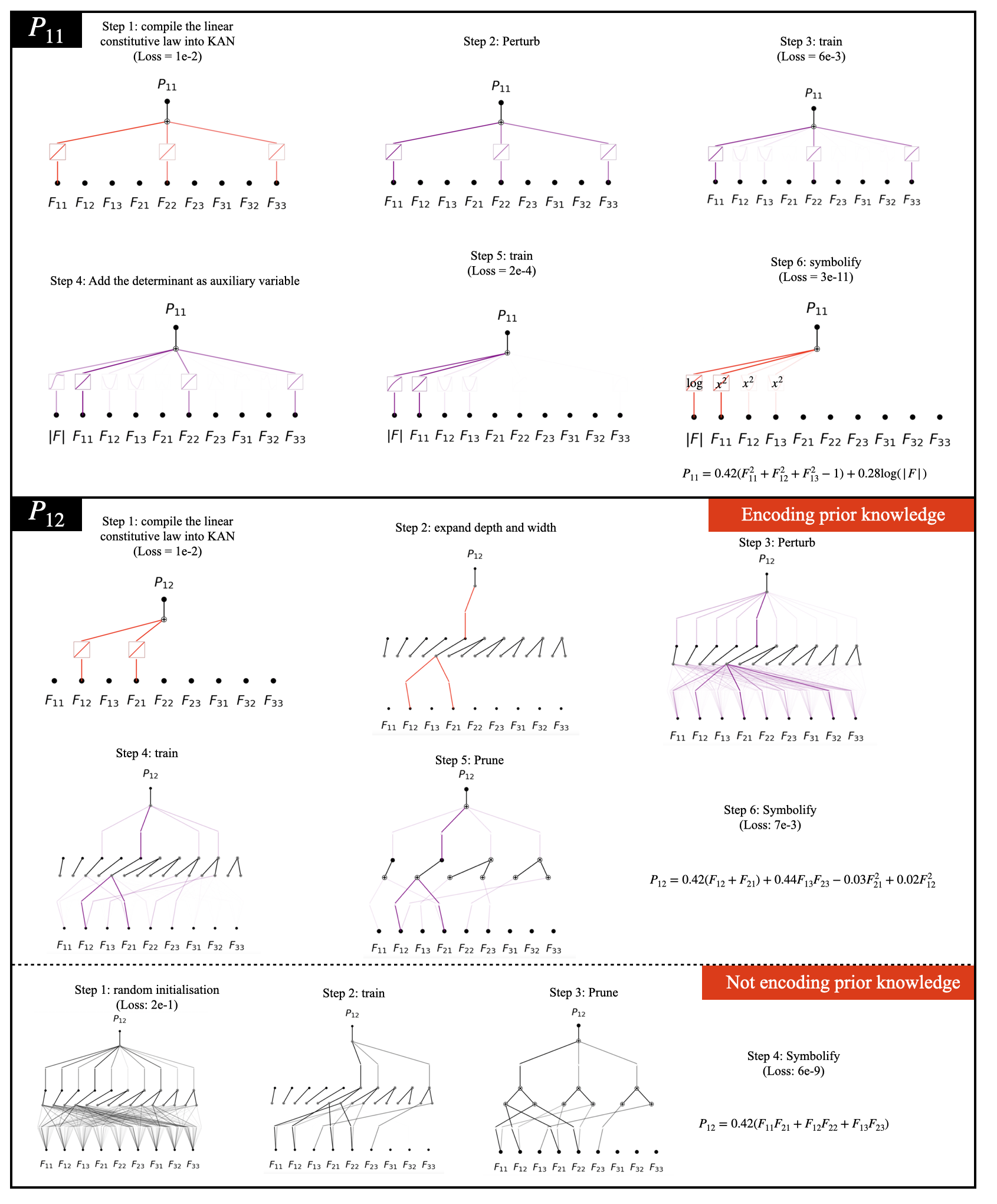}
    \caption{Discovering constitutive laws (relations between the pressure tensor $P$ and the strain tensor $F$) with KANs by interacting with them. Top: predicting the diagonal element $P_{11}$; bottom: predicting the off-diagonal element $P_{12}$. }
    \label{fig:constitutive-law}
\end{figure}

A constitutive law defines the behavior and properties of a material by modeling how it responds to external forces or deformations. One of the simplest forms of constitutive law is Hooke's Law~\cite{hooke2016lectures}, which relates the strain and stress of elastic materials linearly. Constitutive laws encompass a wide range of materials, including elastic materials~\cite{slaughter2012linearized,ogden1997non}, plastic materials~\cite{mises1913mechanik}, and fluids~\cite{batchelor2000introduction}. Traditionally, these laws were derived from first principles based on theoretical and experimental studies~\cite{sifakis2012fem,smith2018stable,arruda1993three,gent1996new}. Recent advancements, however, have introduced data-driven approaches that leverage machine learning to discover and refine these laws from dedicated datasets~\cite{sanchez2020learning,xu2015nonlinear,ma2023learning,ma2024llm}. We follow the standard notations and experimental setups in the elasticity part of NCLaw~\cite{ma2023learning} and define the constitutive law as a parameterized function $\mathcal{E}_\theta(\mat{F})\rightarrow\mat{P}$, where $\mat{F}$ denotes the deformation tensor, $\mat{P}$ the first Piola–Kirchhoff stress tensor, and $\theta$ the parameters in the constitutive law.

Many isotropic materials have linear constitutive laws when deformation is small:
\begin{equation}
      \mat{P}_l = \mu (\mat{F}+\mat{F}^T - 2\mat{I}) + \lambda ({\rm Tr}(\mat{F}) - 3) \mat{I}.
\end{equation}
However, when deformation gets larger, nonlinear effects start to kick in. For example, a Neo-Hookean material has the following constitutive law:
\begin{equation}
      \mat{P} = \mu (\mat{F}\mat{F}^T - \mat{I}) + \lambda {\rm log}({\rm det}(\mat{F}))\mat{I},
\end{equation}
where $\mu$ and $\lambda$ are the so-called Lam\'{e} parameters determined by the so-called Young's modulus $Y$ and Poisson ratio $\nu$ as $\mu = \frac{Y}{2(1+\nu)}, \lambda=\frac{Y\nu}{(1+\nu)(1-2\nu)}$. For simplicity, we choose $Y=1$ and $\nu=0.2$, hence  $\mu=\frac{5}{12}\approx 0.42$ and $\lambda = \frac{5}{18}\approx 0.28$.

Assuming we are working with Neo-Hookean materials, and our goal is to use KANs to predict the $\mat{P}$ tensor from the $\mat{F}$ tensor. Suppose we do not know they are neo-Hookean materials, but we have the prior knowledge that the linear constitutive law is approximately valid for small deformation. Due to symmetries, it suffices to demonstrate that we can accurately predict $P_{11}$ and $P_{12}$ from the 9 matrix elements of $\mat{F}$. We want to compile linear constitutive laws into KANs, which are $P_{11} = 2\mu(F_{11}-1) + \lambda (F_{11}+F_{22}+F_{33}-3)$, and $P_{12} = \mu(F_{12}+F_{21})$. We want to extract Neo-Hookean laws from trained KANs, which are $P_{11} = \mu(F_{11}^2+F_{12}^2+F_{13}^2-1) + \lambda{\rm log}({\rm det}(\mat{F}))$, and $P_{12} = \mu(F_{11}F_{21}+ F_{12}F_{22}+F_{13}F_{23})$.
We generate a synthetic dataset by sampling $F_{ij}$ independently from $U[\delta_{ij}-w, \delta_{ij}+w] (w=0.2)$ and using the Neo-Hookean constitutive law to compute $\mat{P}$. Our interaction with KANs is illustrated in  Figure~\ref{fig:constitutive-law}. In both cases, we successfully figured out the true symbolic formulas in the end, with the aid of some inductive biases. However, the key takeaway is not that we can rediscover the exact symbolic formulas -- given that prior knowledge skews the process -- but rather in real-world scenarios, where the answers are unknown and users can make guesses based on prior knowledge, the pykan package makes it easy to test or incorporate prior knowledge.

{\bf Predicting $P_{11}$} In step 1, we compile the linear constitutive law $P_{11}=2\mu(F_{11}-1)+\lambda(F_{11}+F_{22}+F_{33}-3)$ to a KAN using kanpiler, resulting in a $10^{-2}$ loss. In step 2, we perturb the KAN so that it becomes trainable (indicated by the color change from red to purple; red denotes a purely symbolic part, while purple indicates that both symbolic and spline parts are active). In step 3, we train the perturbed model until convergence, giving a $6\times 10^{-3}$ loss. In step 4, assuming that the determinant is a key auxiliary variable, we use \texttt{expand\_width} (for the KAN) and \texttt{augment\_input} (for the dataset) to include the determinant $|F|$. In step 5, we train the KAN until convergence, giving a $2\times 10^{-4}$ loss. In step 6, we symbolify the KAN to obtain a symbolic formula $P_{11}=0.42(F_{11}^2+F_{12}^2+F_{13}^2-1)+0.28{\rm log}(|F|)$, which achieves a $3\times 10^{-11}$ loss. 

{\bf Predicting $P_{12}$} We experimented with and without encoding the linear constitutive law as prior knowledge. {\bf With prior knowledge}: in step 1, we compile the linear constitutive law to a KAN, resulting in a loss of $10^{-2}$. We then perform a series of operations, including expand (step 2), perturb (step 3), train (step 4), prune (step 5) and finally symbolic (step 6). The influence of prior knowledge is evident, as the final KAN only identifies minor correction terms to the linear constitutive law. The final KAN is symbolified as $P_{12}=0.42(F_{12}+F_{21})+0.44F_{13}F_{23}-0.03F_{21}^2+0.02F_{12}^2$ which yields a $7\times 10^{-3}$ loss, only slightly better than the linear constitutive law. {\bf Without prior knowledge}: in step 1, we randomly initialize the KAN model. In step 2, we train the KAN with regularization. In step 3, we prune the KAN to be a more compact model. In step 4, we symbolify the KAN, yielding $P_{12}=0.42(F_{11}F_{21}+F_{12}F_{22}+F_{13}F_{23})$, which closely matches the exact formula, achieving a $6\times 10^{-9}$ loss. Comparing the two scenarios -- one with and one without prior knowledge -- reveals a surprising outcome: in this example, prior knowledge appears harmful, possibly because the linear constitutive law is probably near a (bad) local minimum which is hard for the model to escape. However, we should probably not randomly extrapolate this conclusion to more complicated tasks and larger networks. For more complicated tasks, finding a local minimum via gradient descent might be challenging enough, making an approximate initial solution desirable. Additionally, larger networks might be sufficiently over-parameterized to eliminate bad local minima, ensuring that all local minima are global and interconnected. 

\section{Related works}

{\bf Kolmogorov-Arnold Networks (KANs)}, inspired by the Kolmogorov-Arnold representation theorem (KART), were recently proposed by Liu et al.~\cite{liu2024kan}. Although the connection between KART and networks has long been deemed irrelevant~\cite{girosi1989representation}, Liu et al. generalized the original two-layer network to arbitrary depths and demonstrated their promise for science-oriented tasks given their accuracy and interpretability. Subsequent research has explored the application of KANs across various domains, including graphs~\cite{bresson2024kagnns,de2024kolmogorov,kiamari2024gkan,zhang2024graphkan}, partial differential equations~\cite{wang2024kolmogorov,shukla2024comprehensive} and operator learning~\cite{abueidda2024deepokan,shukla2024comprehensive,nehma2024leveraging}, tabular data~\cite{poeta2024benchmarking}, time series~\cite{vaca2024kolmogorov,genet2024tkan,xu2024kolmogorov,genet2024temporal}, human activity recognition~\cite{liu2024ikan,liu2024initial},neuroscience~\cite{yang2024endowing,herbozo2024kan}, quantum science~\cite{kundu2024kanqas, li2024coeff,ahmed24graphkan}, computer vision~\cite{cheon2024kolmogorov,azam2024suitability,li2024u,cheon2024demonstrating,seydi2024unveiling,bodner2024convolutional}, kernel learning~\cite{zinage2024dkl}, nuclear physics~\cite{liu2024complexityclaritykolmogorovarnoldnetworks}, electrical engineering~\cite{peng2024predictive}, biology~\cite{pratyush2024calmphoskan}. Liu et al. used B-splines to parameterize 1D functions, and other research have explored various activation functions, including wavelet~\cite{bozorgasl2024wav,seydi2024unveiling}, radial basis function~\cite{li2024kolmogorov}, Fourier series~\cite{xu2024fourierkan}), finite basis~\cite{howard2024finite,ta2024bsrbf}, Jacobi basis functions~\cite{aghaei2024fkan}, polynomial basis functions~\cite{seydi2024exploring}, rational functions~\cite{aghaei2024rkan}. Other techniques for KANs have also been proposed including regularization~\cite{altarabichi2024dropkan}, Kansformer (combining transformer and KAN)~\cite{chen2024sckansformer}, adaptive grid update~\cite{rigas2024adaptive}, federated learning~\cite{zeydan2024f} , Convolutional KANs~\cite{bodner2024convolutional}. There have been ongoing debates regarding whether KANs really outperform other neural networks (especially MLPs) on various domains~\cite{azam2024suitability,cheon2024demonstrating, le2024exploring,shen2024reduced,yu2024kan}, which suggests that while KANs show promise for machine learning tasks, further development is needed to surpass state-of-the-art models. 

{\bf Machine Learning for Physical Laws} A major goal for KANs is to aid in the discovery of new physical laws from data. Previous research has shown that machine learning can be used to learn various types of physical laws, including equations of motion~\cite{wu2019toward, brunton2016discovering, lemos2023rediscovering, cranmer2020discovering}, conservation laws~\cite{aipoincare, aipoincare2, sid, lu2023discovering, ha2021discovering, sebastian-cl}, symmetries~\cite{krippendorf2020detecting, liu2022hiddensymmetry, yang2023generative}, phase transitions~\cite{wetzel2017unsupervised,carrasquilla2017machine}, Lagrangian and Hamiltonian~\cite{cranmer2020lagrangian, greydanus2019hamiltonian}, and symbolic regression~\cite{cranmer2023interpretable, martius2016extrapolation, dugan2020occamnet, schmidt2009distilling}, etc. However, making neural networks interpretable often requires domain-specific knowledge, limiting their generality. We hope that KANs will evolve into universal foundation models for physical discoveries. 

{\bf Mechanistic Interpretability} seeks to understand how neural networks operate in a fundamental level~\cite{cunningham2023sparse,meng2022locating,wang2023interpretability,elhage2022toy, nanda2023progress, zhong2023the,liu2023seeing,elhage2022solu,li2023emergent,engels2024not}. Some research in this area focuses on designing models that are inherently interpretable~\cite{elhage2022solu} or proposing training methods that explicitly promote interpretability~\cite{liu2023seeing}. KANs fall into this category since the Kolmogorov-Arnold theorem decomposes a high-dimensional function into a collection of 1D functions, which are significantly easier to interpret than high-dimensional functions.

\section{Discussion}

\begin{figure}
    \centering
    \includegraphics[width=1.0\linewidth]{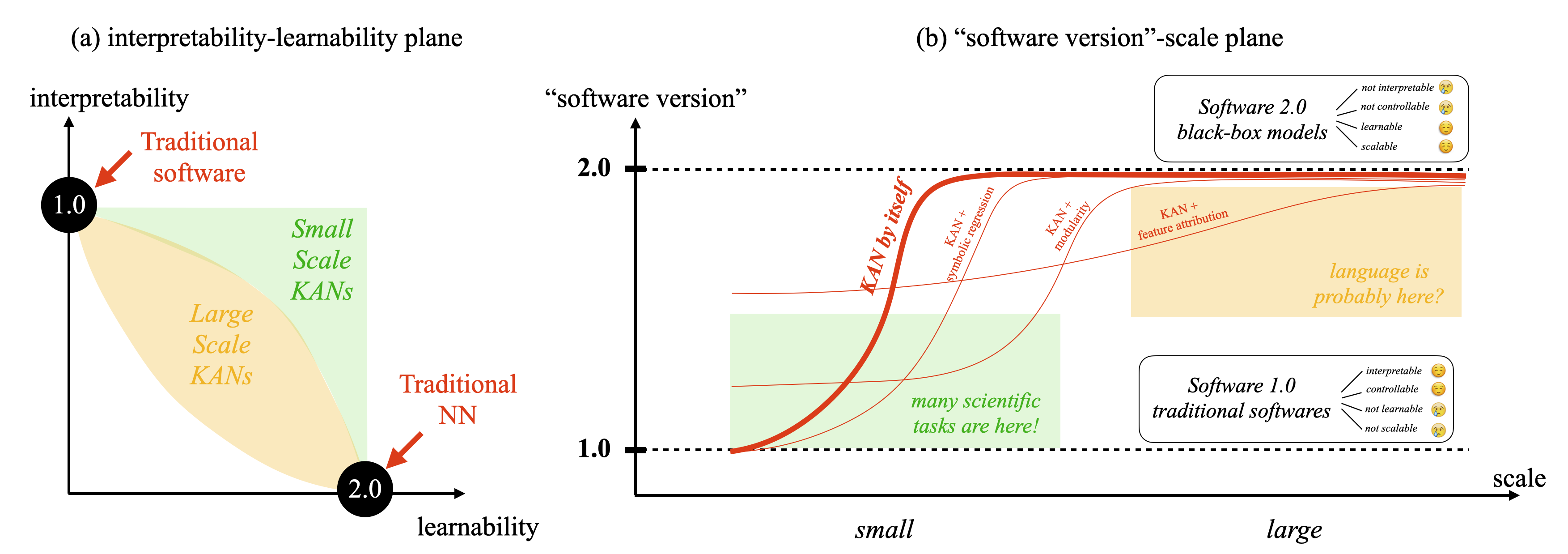}
    \caption{KAN interpolates between software 1.0 and 2.0. (a) KANs strike a balance between interpretability (software 1.0) and learnability (software 2.0). (b) KANs' Pareto frontier on the interpretability-scale plane. The amount of interpretation we can get from KANs depends on problem scales and interpretability methods.}
    \label{fig:software}
\end{figure}

{\bf KAN interpolates between software 1.0 and 2.0} The key difference between Kolmogorov-Arnold Networks (KANs) and other neural networks (\textit{software 2.0}, a term coined by Andrej Karpathy) lies in their greater interpretability, which allows for manipulation by users, similar to traditional software (\textit{software 1.0}). However, KANs are not entirely traditional software, as they (1) learnability (good), enabling them to learn new things from data, and (2) reduced interpretability (bad) as they become less interpretable and controllable as the network scales increase. In Figure~\ref{fig:software} (a) visualizes the position of software 1.0, software 2.0, and KANs on the interpretability-learnability plane, illustrating how KANs can balance the trade-offs between these two paradigms.  
The goal of this paper is to propose various tools that make KANs more like software 1.0, while leveraging the learnability of software 2.0.

{\bf Efficiency improvement} The original pykan package~\cite{liu2024kan} was poor in efficiency. We have incorporated a few techniques to improve its efficiency.
\begin{enumerate}
    \item Efficient splines evaluations. Inspired by  Efficient KAN~\cite{Blealtan}, we have optimized spline evaluations by avoiding unnecessary input expansions. For a KAN with $L$ layers, $N$ neurons per layer, and grid size $G$, memory usage has been reduced from $O(LN^2G)$ to $O(LNG)$.
    \item Enabling the symbolic branch only when needed. A KAN layer contains both a spline branch and a symbolic branch. The symbolic branch is much more time-consuming than the spline branch since it cannot be parallelized (disastrous double loops are needed). However, in many applications, the symbolic branch is unnecessary, so we can skip it when possible, significantly reducing runtime, especially when the network is large. 
    \item Saving intermediate activations only when needed. To plot KAN diagrams, intermediate activations must be saved. Initially, activations were saved by default, leading to slower runtime and excessive memory usage. We now save intermediate activations only when needed (e.g., for plotting or applying regularizations in training). Users can enable these efficiency improvements with a single line: \texttt{model.speed()}. 
    \item GPU acceleration. Initially, all models were run on CPUs due to the small-scale nature of the problems. We have now made the model GPU-compatible~\footnote{Models can be trained on GPUs, but not all functionalities already supported GPU.}. For example, training a [4,100,100,100,1] with Adam for 100 steps used to take an entire day on a CPU (before implementing 1, 2, 3), but now takes 20 seconds on a CPU and less than one second on a GPU. However, KANs still lag behind MLPs in efficiency, especially at large scales. The community has been working towards benchmarking and improving KAN's efficiency and the efficiency gap has been significantly reduced~\cite{JerryMaster}.
\end{enumerate}
Since the objective of this paper is to make KANs more like software 1.0, when facing trade-offs between 1.0 (being interactive and versatile) and 2.0 (being efficient and specific), we prioritize interactivity and versatility over efficiency. For example, we store cached data within models (which consumes additional memory), so users can simply call \texttt{model.plot()} to generate a KAN diagram without manually doing a forward pass to collect data.

{\bf Interpretability} Although the learnable univariate functions in KANs are more interpretable than weight matrices in MLPs, scalability remains a challenge. As KAN models scale up, even if all spline functions are interpretable individually, it becomes increasingly difficult to manage the combined output of these 1D functions. Consequently, a KAN may only remain interpretable when the network scale is relatively small (Figure~\ref{fig:software} (b), thick red line). It is important to note that interpretability depends on both intrinsic factors (related to the model itself) and extrinsic factors (related to interpretability methods). Advanced interpretability methods should be able to handle interpretability at various levels. For example, by interpreting KANs with symbolic regression, modularity discovery and feature attribution (Figure~\ref{fig:software} (b), thin red lines), the Pareto Frontier of interpretability versus scale extends beyond what a KAN alone can achieve. A promising direction for future research is to develop more advanced interpretability methods that can further push the current Pareto Frontiers.

{\bf Future work} This paper introduces a framework that integrates KANs with scientific knowledge, focusing primarily on small-scale, physics-related examples. Moving forward, two promising directions include applying this framework to larger-scale problems and extending it to other scientific disciplines beyond physics.



\section*{Acknowledgement}
We would like to thank Yizhou Liu, Di Luo, Akash Kundu and many GitHub users for fruitful discussion and constructive suggestions. We extend special thanks to GitHub user Blealtan for making public their awesome work on making KANs efficient. Z.L. and M.T. are supported by IAIFI through NSF grant PHY-2019786.

\bibliographystyle{abbrv}
\bibliography{ref}

\end{document}